%% file: main.tex
\documentclass[conference]{IEEEtran}
\IEEEoverridecommandlockouts
\usepackage{cite}
\usepackage{amsmath,amssymb,amsfonts}
\usepackage{algorithm}
\usepackage{algpseudocode}
\usepackage{graphicx}
\usepackage{textcomp}
\usepackage{subfiles}
\usepackage{subcaption}
\usepackage{multicol}
\usepackage{multirow}
\usepackage{booktabs}
\usepackage[table,xcdraw, dvipsnames]{xcolor}
\usepackage{url}
\usepackage{todonotes}
\usepackage[english]{babel}
\usepackage{amsthm}
\usepackage{amsmath}
\usepackage[utf8]{inputenc}
\usepackage[english]{babel}

\theoremstyle{definition}
\newtheorem{definition}{Definition}[section]
\newtheorem{exmp}{Example}[section]
\usepackage[multiple]{footmisc}

\usepackage{graphicx,pifont}

\def\BibTeX{{\rm B\kern-.05em{\sc i\kern-.025em b}\kern-.08em
    T\kern-.1667em\lower.7ex\hbox{E}\kern-.125emX}}

\newcommand{\para}[1]{\vspace{1mm}\noindent\textbf{#1}.} 

\newcounter{Asterios}
\stepcounter{Asterios}

\newcounter{Ale}
\stepcounter{Ale}

\newcounter{RihanNOC}
\stepcounter{RihanNOC}

\newcounter{ziyu}
\stepcounter{ziyu}

\newcounter{hilco}
\stepcounter{hilco}

\newcommand{\revision}[1]{\textcolor{black}{#1}}

\newcommand{\revisiontwo}[1]{\textcolor{black}{#1}}

\newcommand{\revisionthree}[1]{\textcolor{black}{#1}}

\newcommand{\revisionfour}[1]{\textcolor{black}{#1}}

\begin{document}
\pagenumbering{Roman}

\bstctlcite{MyBSTcontrol} 

\title{Model Selection with Model Zoo via Graph Learning\\
}

\author{
    \IEEEauthorblockN{
     Ziyu Li \qquad  Hilco van der Wilk \qquad Danning Zhan \\ \qquad Megha Khosla \qquad Alessandro Bozzon \qquad Rihan Hai} 
    \IEEEauthorblockA{Delft University of Technology, The Netherlands}
    \IEEEauthorblockA{z.li-14,\{initials.lastname\}@tudelft.nl}
}

\maketitle

\input{sections/0_abstract}
\input{sections/1_intro}

\input{sections/problem}

\input{sections/3_method}

\input{sections/4_evaluation}

\input{sections/2_related}

\input{sections/5_conclusion}
\bibliographystyle{IEEEtran}
\bibliography{IEEEabrv, main, references}

\newpage
\input{sections/appendix}

\end{document}

%% file: sections/0_abstract.tex
\begin{abstract}
Pre-trained deep learning (DL) models are increasingly accessible in public repositories, i.e., model zoos. 
Given a new prediction task, finding the best model to fine-tune can be computationally intensive and costly, especially when the number of pre-trained models is large.
Selecting the right pre-trained models is crucial, yet complicated by the diversity of models from various model families (like ResNet, Vit, Swin) and the hidden relationships between models and datasets. 
Existing methods, which utilize basic information from models and datasets to compute scores indicating model performance on target datasets, overlook the intrinsic relationships, limiting their effectiveness in model selection. 
In this study, we introduce TransferGraph, a novel framework that reformulates model selection as a graph learning problem. TransferGraph constructs a graph using extensive metadata extracted from models and datasets, while capturing their inherent relationships. 
Through comprehensive experiments across \revisiontwo{16} real datasets, \revisiontwo{both images and texts}, we demonstrate TransferGraph's effectiveness in capturing essential model-dataset relationships, yielding up to a 32\% improvement in correlation between predicted performance and the actual fine-tuning results compared to the state-of-the-art methods.

\end{abstract}

%% file: sections/1_intro.tex
\section{Introduction}
\label{sec:intro}

Deep learning has been widely used in handling \revisiontwo{unstructured} data,  including tasks related to image \revisiontwo{and text classification}. The paradigm of \emph{first pre-training, then fine-tuning} has become the de facto of applying deep learning in practice. Pre-training is the phase of training a neural network on a large, diverse dataset, typically drawn from a general domain, e.g., ImageNet~\cite{deng_imagenet_2009}. Subsequently, the fine-tuning step refines the model for a specific task, often a smaller target dataset. 
This two-step process leverages the general knowledge acquired during pre-training, facilitating effective adaptation to a narrower and more specialized context.
The general representations learned during pre-training, speed up model convergence during fine-tuning and help reduce the risk of over-fitting.

Today, many pre-trained models are available in public online platforms, e.g., HuggingFace\footnote{\url{https://huggingface.co/}}, TensorFlow Hub\footnote{\url{https://www.tensorflow.org/}}, and PyTorch Hub\footnote{\url{https://pytorch.org/hub/}}.
Such repositories of pre-trained models are referred to as \emph{model zoos}.
Model zoos have been widely adopted in recent years, as they offer convenient access to a collection of pre-trained models, including cutting-edge deep learning architectures.
This lowers the expertise barrier, enabling non-expert individuals to apply complex deep learning models in their applications. 
Utilizing a model zoo for fine-tuning facilitates the adaptation across a wide range of target datasets, which have varying quantities of training data~\cite{deshpande2021linearized}. 
In addition, by fine-tuning pre-trained models from the model zoo, machine learning practitioners can bypass the need for training from scratch—a resource-intensive process, resulting in significant savings in both development time and computational resources. 

However, it is a non-trivial task to pick the \emph{right} pre-trained models as the starting point of fine-tuning, which has a substantial impact on the effectiveness of the fine-tuning results~\cite{deshpande2021linearized}.
A straightforward solution is to fine-tune all the relevant pre-trained models, which is computationally expensive, and sometimes infeasible in practice.
For instance, there are 7411 models for image classification tasks in the HuggingFace repository and 900 variations on TensorFlow Hub.
It took 1178 hours of GPU time to fine-tune all the 185 models in our model zoo on a single dataset.

\begin{figure}[t]
\centering
	\includegraphics[width=0.47\textwidth]{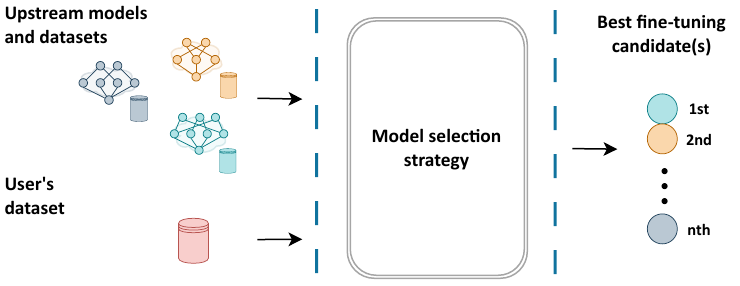}
		\caption{Illustration of the model selection problem setting.}
		\label{fig:problem-setting}
\end{figure}

 The practical choice is to identify pre-trained models that exhibit promising performance even without fine-tuning, i.e., \emph{model selection}. 
 As in Figure~\ref{fig:problem-setting}, given a target dataset and several pre-trained models over existing datasets, model selection aims to rank and select optimal candidates from the model zoo to perform fine-tuning. 
 Different strategies may yield disparate rankings of the candidates.


\begin{figure}[t]
\centering
	\includegraphics[width=0.35\textwidth]{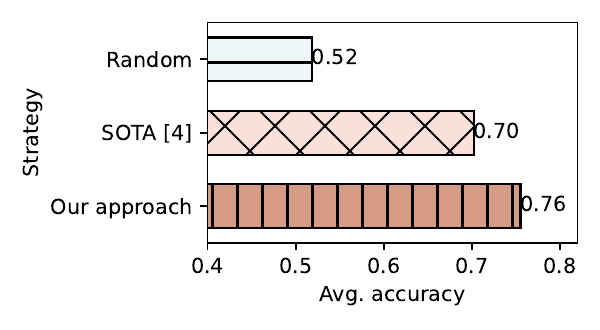}
		\caption{Average fine-tuned accuracy of the top 5 selected models compared between random selection strategy and our proposed solution learning from a graph along with metadata (example dataset: stanfordcars~\cite{krause_3d_2013}).}
		\label{fig:random_selection}
\end{figure}

A naive approach is to randomly select models for fine-tuning. 
This random selection strategy may suffice when pre-trained models all have similar fine-tuning accuracy. However, in the more general case where the performance of models varies, the random strategy is ineffective. 
In Figure~\ref{fig:random_selection}, we report the results of the average accuracy of the top five models selected through diverse strategies (full results in Section~\ref{sec:evaluation}). \texttt{Random} denotes a random selection strategy, which only achieved an unsatisfying accuracy value of 0.52. 

Existing studies~\cite{you_logme_2021, tran_transferability_2019, nguyen_leep_2020,  bolya_scalable_2021, huang_frustratingly_2022} mainly focus on extracting information about the pre-trained models and datasets, and mapping model features to the target dataset labels to measure the model \textit{transferability}. 
The efficacy of features is expected to diminish as the source dataset (training dataset of the pre-trained model) and target dataset become less similar~\cite{yosinski2014transferable}.
\revision{Another approach, exemplified by Amazon LR \cite{li_guided_2023}, learns the pattern of model performance by using metadata (e.g., model architecture, data size) to train a regression model.}
\revisiontwo{The mechanism of the previous methods is limited to applying model representations extracted from the learned parameters or features constructed from metadata, overlooking the deeper connections inherent among models and datasets. }
  
\revisiontwo{
Our work advances beyond existing studies by incorporating the prior knowledge of fine-tuning and transferability scores (e.g., LogME~\cite{you_logme_2021}), representing this information through weighted connections between models and datasets.}
We borrow the inspiration from data management systems for data repositories, such as data lakes \cite{nargesian2019data, terrizzano2015data, 10107808}. 
For managing a collection of datasets,  a common approach is to structure these datasets as graphs \cite{DBLP:conf/icde/FernandezAKYMS18, zhang2020finding, nargesian2020organizing}. This involves representing tables as nodes and their relationships as edges. For instance, an edge can indicate that two tables are semantically similar \cite{DBLP:conf/icde/FernandezAKYMS18}. 
For the model selection problem, rich relationships exist not only between models and datasets, but also among datasets themselves. 
\revisiontwo{Our approach leverages the additional information on relationships informed by fine-tuning and transferability scores, and dataset similarity.}



We reformulate the challenge of model selection as a graph link prediction problem. 
We propose \emph{TransferGraph}\footnote{\revisiontwo{Code is available at \url{https://github.com/TransferGraph/transfergraph} and HuggingFace organization at \url{https://huggingface.co/TransferGraph}}}, which explores how the relationships among \emph{dataset-dataset} and \emph{dataset-model} can facilitate more effective model selection, 
offering a structured and intuitive method to navigate and understand these complex relationships.
To represent and analyze these intricate relationships, we represent them using graph structures. 
We show that TransferGraph is able to identify suitable pre-trained models for the target dataset by exploiting graph features learned from the graph structure and along with other metadata information (e.g., model architecture, data size). 
As shown in Figure~\ref{fig:random_selection}, TransferGraph outperforms the state-of-the-art method \cite{you_logme_2021} with a notable improvement in fine-tuning accuracy.

\para{Contributions} We summarize our contributions as follows. 
\begin{itemize}
    \item We reformulate the model selection problem as a graph learning problem.
    \item Different from existing works, which only take into account the dataset labels or solely information about models or datasets, we exploit the metadata of both models and datasets and further learn the inherent relationships between the artifacts by learning a graph.
    \item We propose a framework that tackles the model selection problem via graph learning. The framework consists of end-to-end processes, from feature collection and graph learning to model performance prediction.
    \item Extensive experiments are conducted to evaluate the validity of our graph-based model selection strategy. 
    Our findings demonstrate that applying a graph learning method exhibits a strong correlation with actual fine-tuning accuracy across image and text classification tasks.
\end{itemize}

%% file: sections/problem.tex
\section{Background and problem definitions}
\label{sec:method}



It is a challenging task to 
select the \emph{right} models for fine-tuning, especially given
the abundant pre-trained models in the model zoo. 
In this section, we explain existing model selection strategies and identify their limitations.

\subsection{Model selection strategies}
 
Previous model selection strategies consider features of models and datasets from different perspectives.
They mainly differ in the information of datasets and pre-trained models they use. Some works like SHiFt \cite{renggli_shift_2022} have developed systems which combine these approaches, while also taking user inputs such as budget constraints into account. 
We categorize these model selection strategies based on the features they employ to rank the models for model selection.


\para{Task-similarity-based model selection} Early model selection methods use the similarity of the source and target tasks to measure the transferability of a model. 
When the target task is similar to source task, a model with good performance on the source task is likely to have good fine-tuning performance \cite{wang_characterizing_2019}. Methods in this group include EMD \cite{cui_large_2018} and NCE~\cite{tran_transferability_2019}, Task2Vec \cite{achille_task2vec_2019,tran_transferability_2019}. 
To obtain the similarity of the source and target tasks, EMD~\cite{cui_large_2018} and NCE~\cite{tran_transferability_2019} compare source and target task features and labels. Task2Vec \cite{achille_task2vec_2019} embeds tasks as vectors using a single probe model and computes their pairwise distances as transferability scores.


\para{Feature-based model selection} More recent approaches leverage the target task specific features, which are extracted by 
 executing a forward pass of the target task on each pre-trained model (model inference on the target dataset).  Methods in this group include LEEP \cite{nguyen_leep_2020}, LogME \cite{you_logme_2021}, PARC \cite{bolya_scalable_2021} and TransRate \cite{huang_frustratingly_2022}.
These approaches circumvent the need for fine-tuning. However, as the number of pre-trained models in a model zoo grows, it becomes inefficient to perform a forward pass over all pre-trained models, even infeasible. Moreover, methods in this group overlook basic features of both the target dataset (like the number of samples and labels) and the pre-trained model (such as input size and architecture), which are crucial for fine-tuning efficiency. 
For instance, a mismatch in input size or the number of classes, leads to significant deterioration in fine-tuning performance\cite{li_guided_2023}. 

\para{Learning-based model selection} The third group of approaches \cite{ li_guided_2023} trains a simple linear model, i.e., a linear regression model, to predict model performance and recommend pre-trained models given a new target task. The used features extracted from models
or \textit{metadata} of the target dataset and pre-trained models. 
The state-of-the-art approach, Amazon LR \cite{li_guided_2023}, employs only basic metadata of the target dataset and pre-trained models. It achieved competitive results when learning a linear regression model. The authors suggest that incorporating additional features could further improve this method. 


\subsection{Limitations and Challenges}
\label{ssec:limit}
We summarize the limitations of existing model selection strategies and outline the challenges to tackle them.


\subsubsection{Overlooking the heterogeneity of model zoo} 

A model zoo may encompass heterogeneous models and datasets.
Models within this context can exhibit differences in the pre-trained domain, architecture, and hyperparameter settings.
At the same time, datasets vary in terms of the tasks they address and the distribution of their data.
Predicting the performance and capability of models is challenging, given that the models are trained differently, and the inductive biases of models are different. 
Featured-based model selection strategies usually use the model as feature extractor, or assumes the fine-tuning process does not change the backbone weights much~\cite{bolya_scalable_2021,deshpande_linearized_2021}. 
However, such an assumption does not hold in practice.

Prior studies~\cite{deshpande_linearized_2021,huang_frustratingly_2022} have often restricted model architectures to certain categories, e.g., ResNet, MobileNet or DenseNet.
In LogMe~\cite{you_logme_2021}, only models pre-training on the same source dataset (e.g., ImageNet) are included.
However, the optimal architecture or Pareto-optimal models are usually task-dependent, relying on the inductive bias of the model and the dataset properties~\cite{li_guided_2023}. 
Fine-tuning with a model zoo helps transfer to a diverse set of target tasks with different downstream datasets.
Due to the diversity of the model characteristics, e.g., architecture family, pre-trained domain, and hyperparameter settings, it is even more challenging to identify suitable candidate models for the downstream task.

\subsubsection{Insufficient feature coverage} Feature-based model selection strategies~\cite{deshpande_linearized_2021,bolya_scalable_2021} often rely on the model features and the target dataset labels.
They assume that models can generalize better if the features extracted by the model are similar and labels are similar.
However, such approaches struggle to accurately predict top-performing models for target datasets significantly different from source datasets used for pre-training~\cite{li_guided_2023}. 
Conversely, incorporating simple prior knowledge, such as dataset characteristics, is proven to help predict the model performance on downstream datasets~\cite{li_guided_2023}. 
Learning from basic metadata of models and datasets is beneficial yet limited due to its coarse-grained nature.
It often overlooks the intricate relationships between models and datasets. 
Therefore, a central challenge lies in identifying and utilizing such inherent relationships for more effective model selection.

\vspace{-2mm}
\section{Model selection as a graph learning problem}
\label{sec:graph} 
We first explain the problem setting of model selection.
To tackle the challenges in Sec.~\ref{ssec:limit}, we propose transforming the model selection problem into a graph learning problem.  

\vspace{-2mm}
\subsection{Problem definition}

Consider a set of models, denoted as $M = \{m_1, ... m_{N}\}$, and a collection of datasets, represented as $\chi = \{d_1, ... d_{K}\}$. 
We denote the actual fine-tuning accuracy as $T_{i, j}$, with respect to the model $m_{i}$ and the target dataset $d_{j}$, where $m_{i} \in M$, and $d_{j} \in \chi$. 
Given a pre-trained model $m_{i}$, we are interested in predicting a score $S_{ij}$ which approximates its fine-tuning accuracy on the target dataset $d_{j}$. 
\begin{exmp}
\label{exmp:model_selection} 
Consider two models $m_{1}$ and $m_{2}$, and two target datasets $d_{1}$ and $d_{2}$. 
The predicted scores of the models on each dataset can be presented by a matrix $ S = \begin{pmatrix}
    0.6 & 0.8\\
    0.7 & 0.3
\end{pmatrix}$. 
For dataset $d_1$, the predicted score $S_{11}$ of $m_1$ on $d_1$ is 0.6, and $S_{21}$ is 0.8. 
It indicates that $m_2$ is predicted to have a better fine-tuning performance than $m_1$.
Whereas it is a different case on the dataset $d_2$, with $S_{12}$=0.7 higher than $S_{22}$=0.3. 
\end{exmp}

The predicted score should be a good approximation of the actual fine-tuning results and exhibit a strong correlation with the target dataset.
Such an alignment would enable the predicted score to be a reliable indicator of fine-tuning performance on the target dataset, allowing for the effective ranking of pre-trained models. 


To measure the effectiveness of the model selection score, we use \emph{Pearson's correlation coefficient}, following the common practice~\cite{li_guided_2023}.
We use $\tau \in [-1,1]$ to represent the Pearson's correlation.
Given $N$ paired data $\{(m_1,d_t),(m_2,d_t),...(m_N,d_t)\}$, $\tau$ is defined as:

\begin{equation}
\label{eq::ine-tuning}
    \tau = \frac{\sum_{i=1}^{N} (T_{i} - \bar{T})(S_{i} - \bar{S})}{\sqrt{\sum_{i=1}^{N} (T_{i} - \bar{T})^{2} \sum_{i=1}^{N} (S_{i} - \bar{S})^{2}}}
\end{equation}

The goal is to maximize the correlation between the predicted scores and the model performance.
An absolute value of 1 implies that a $S$ perfectly aligns with the trend of $T$ with all data points lying on a line. 

\begin{table}
    \begin{center}
        \caption{Notation definitions}
            \label{tab:graph}
    \begin{tabular}{ c|c } 
     \hline
      \textbf{Notation} & \textbf{Definition}\\ 
     \hline\hline
     
      $S_{i,j}$ & Predicted transferability score of $m_i$ on $d_j$ \\\hline
      $T_{i,j}$  & Fine-tuning performance of $m_i$ on $d_j$ \\\hline
      $\chi, d_j$     & Set of datasets collection and dataset $j$ \\\hline
      $M$, $m_i$   & Model collection and model $i$ \\\hline
      $\phi$       & Dataset similarity \\\hline
      $G$       & Graph \\\hline
      $E$       & Edge of Graph \\\hline
      $V$       & Vertex / Node \\\hline
      $L$       & Edge labels \\\hline
      $\tau$    & Pearson's Correlation \\\hline
      $f_{G}()$ & Function over the graph \\\hline
      $W^{(k)}$ & Weights of the graph \\\hline
      $Q^{(k)}$ & Operation that allows aggregation afterwards \\\hline
      \\[-1em]
      $ \bar{X} $ & Mean of the variable X \\\hline
      $\hat{X}$ & Prediction for the variable X \\\hline
      $F()$     & Learned function to predict the model performance \\\hline
    \end{tabular}
    \end{center}
    \vspace{-0.4cm}
\end{table}

\subsection{Convert model selection to graph learning}

In this work, we formulate the model selection problem as a graph learning process, which maximizes the Pearson correlation $\tau$, between the predicted score $S_{ij}$ and the actual fine-tuning accuracy $T_{ij}$.

\begin{definition}[Graph] \label{def::graph}
We denote a graph as $G = (V, E)$ where
$V$ is the set of vertices/nodes and
$E \subseteq V\times V$ denotes the set of edges that connect the vertices in $V$. 
\end{definition}


In our setting, a vertex either represents a dataset or a model.
Given a set of datasets $\chi$ = \{$d_1$, $d_2$, ..., $d_{K}$\}, and a set of pre-trained deep learning models $M = \{m_1 , m_2$, $..., m_N \}$, we build our vertex set as $V = \chi \cup M$.
Here $K = |\chi|$ is the number of datasets, and $N = |M|$ is the number of models.


In our graph, we construct three types of edges, depending on the nodes connected by the edges. 
The first type links dataset nodes, utilizing calculated dataset similarity for connection. The second type forms connections between a model and a dataset, representing existing transferability scores, e.g., LogME~\cite{you_logme_2021}, PARC \cite{bolya_scalable_2021}. 
The third edge type also connects a model and a dataset, and comes from the training history of each model on each dataset, such as the pre-trained performance and fine-tuning performance.
In the graph, instead of having the binary adjacency matrix, the respective scores will be used as the weights of the adjacency matrix.
Instead of having a fully connected graph, a pruning threshold will be used to decide the existence of the edges.
\begin{exmp}
\label{exmp:as_a_graph} 
If we have two datasets $d_{1}, d_{2}$ and two models $m_{1}, m_{2}$ then we can form the graph with edge sets $E_{G} = \{ (d_{1}, d_{2}), (d_{1}, m_{2}), (d_{2}, m_{2}), (d_{2}, m_{1})\}$.
Each of these edges will have a value, as per a weighted adjacency matrix, the value for $(d_{1}, d_{2})$ will be the similarity score $\phi_{d_1,d_2}$ between the datasets. 
The value between edges $\{ (d_{1}, m_{1}), (d_{1}, m_{2}), (d_{2}, m_{2}), (d_{2}, m_{1})\}$ will be the training performance of the model on the corresponding dataset.
These values can be taken from Example \ref{exmp:model_selection} and constructing the graph using those values.
\end{exmp}

In this work, we are interested in exploiting the inherent relationships between models and datasets for the model selection problem in a model zoo. 
Borrowing the concept from data lake management, we represent the model and dataset relationships in a graph and learn the graph structure by performing a link prediction task.

\para{Link prediction} 
We extend Definition~\ref{def::graph} to $G=(V,E,L)$, by adding the set of labels or representations of each edge, denoted as $L$. 
The goal of link prediction is to learn a predictive model that assigns a score to pairs of nodes $(u,v)$, indicating the likelihood of an edge existing between them.

\begin{figure}[t]
\centering
	\includegraphics[width=0.5\textwidth]{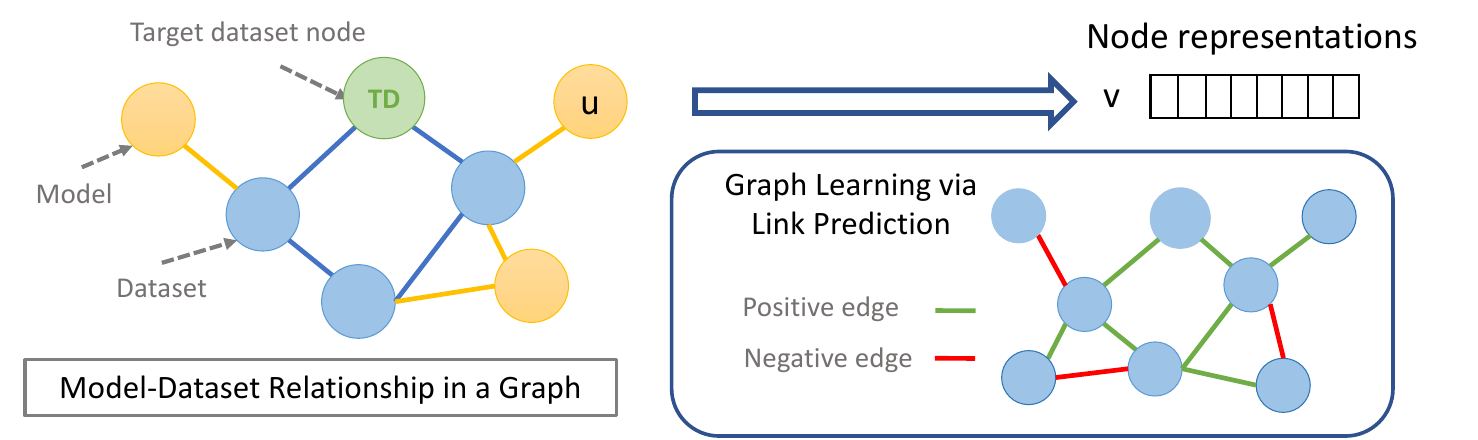}
		\caption{Link prediction in the context of model selection}
		\label{fig:link-prediction}
  \vspace{-0.5cm}
\end{figure}

We aim to identify the models that have high performance on the datasets. We can specify the positive edges with models receiving high performance in the training history. 
\revisionthree{We regard the link prediction task as a classification problem, positive edges as 1 and negative edges as 0.} 
Thus edges of models performing well on a dataset \revision{are regarded as positive edges, and vice versa.}
Illustration is shown in Figure~\ref{fig:link-prediction}.



We formulate the model selection problem through a learned function over the graph, represented by the following formula.
\begin{equation}
    \label{formula::ms}
    \Hat{{T}}_{i,j} = F(f_{G}(m_{i}), f_{G}(d_{j})),
\end{equation}

In Equation \ref{formula::ms}, we use the graph learners $f_{G}$ to learn the set of labels $L$ for the link prediction task on our constructed graph.
$f_G(m_{i})$ obtains the vertex embeddings of $m_i$, and $f_G(d_j)$ obtaining the vertex embeddings of $d_j$.
$F$ denotes the prediction model that maps from the model and data representations to the fine-tuning results.
The prediction model is trained on the training history.

We reformulate the model selection problem with a model zoo as a graph link prediction problem. 
In what follows, we will introduce the information needed to tackle the problem in our proposed graph-learning-based strategy.




%% file: sections/3_method.tex
\section{Data collection: Metadata and Features}
\label{sec:features}

Extensive research~\cite{nguyen_leep_2020,you_logme_2021,bolya_scalable_2021} has been conducted to investigate the relationship between the model features and the target dataset labels.
Yet, the metadata of models and datasets are often neglected.
Though simple and coarse-grained, such metadata are of great value to specify the characteristics of the models and datasets in some sense, and prove to be useful for predicting the fine-tuning performance~\cite{li_guided_2023}.
Below, we will introduce the main metadata and features considered. 

\subsection{Metadata as features}
In the following subsections, we present the considered metadata of both models and datasets. 

\subsubsection{Metadata of datasets}
\label{sec:dataset-metadata}

The metadata of datasets can be indicators of the fine-tuning difficulty.
The properties of a dataset can affect a model's performance.
For example, a dataset with many classes is more difficult to learn than a dataset with binary classes. 
We do not exclude the information from the pre-trained model, as in most feature-based model selection strategies.
We consider the metadata of both source and target datasets for model selection.

\para{Number of data samples}
A small dataset contains less information and is likely easier to learn. 
In contrast, a large dataset with more diverse features may require a more complex model to learn to obtain good performance.

\para{Number of label classes}
A multi-label classification problem is more challenging than binary classification and may require more data samples to learn.

\subsubsection{Metadata of models}
\label{sssec:metadata_model}
The metadata of models reveals their learning capability from a certain perspective. A model with more parameters may capture more generalized features.
Models with different architectures may have varying inductive biases for different datasets.

\para{Input shape} 
More information can be captured with a larger input shape, e.g., a higher-resolution image.

\para{Architecture}
The architecture of a model plays an important role in determining how well a model can capture complex patterns in a dataset.
A more complex architecture, e.g., ResNet~\cite{he2016identity}, Inception~\cite{dosovitskiy_image_2021}, might be more suitable to learn more complex and larger inputs than e.g., LeNet~\cite{lecun_gradient-based_1998}.

\para{Pre-trained dataset}
The source data quality significantly impacts the learned features and knowledge that a model can capture.
A model trained on a large dataset with diverse data may have more generalized ability than one trained on a small and biased dataset.

\para{Model performance}
The performance identifies the capability of a model. 
For example, when two models are trained on the same dataset, the model with higher accuracy indicates that it has better knowledge of the dataset and may be adaptable to new datasets.

\para{Number of parameters} 
A bigger model with more parameters can capture more generalized features from a large dataset. 
Compared to an SVM model, a more complex model (ResNet) can perform better in image classification on ImageNet. 

\para{Memory consumption} 
The memory consumption of a model is correlated to the number of parameters.
It is another indicator of the complexity of a model.

This work does not include all the metadata mentioned in Amazon LR~\cite{li_guided_2023}.
Some metadata included in Amazon LR needs further computation to obtain, e.g., dataset difficulty.
The metadata mentioned above are more accessible to obtain. 
In addition, we include some other features, e.g., models' pre-trained performance compared to Amazon LR.
We find that even with the simple metadata, the model selection strategies can make good predictions on the model performance.

\subsection{Dataset Features}

Together with the metadata, we capture the dataset features in the feature collection stage.
Similar to feature-based model selection strategies, which acquire features by executing a forward pass over all candidate models on the target dataset, we can capture dataset representations through a comparable approach. By utilizing a reference ML model, referred to as a \emph{probe network}, for inference on datasets as the initial step.
We acknowledge that the probe network exhibits varied performance on different datasets, resulting in distinct embeddings within a vector space. 
We expect these embeddings to unveil the distinctive characteristics of the datasets, and the distance between embeddings captures the semantic similarities between the datasets.

\subsubsection{Dataset representations} 
Prior studies, including Domain Similarity~\cite{cui_large_2018}, Task2Vec~\cite{achille_task2vec_2019} and Taskonomy~\cite{zamir_taskonomy_2018}, focus on learning dataset representations within the realm of transfer learning. 
We adopt \textit{Domain Similarity} to extract the embeddings for dataset representations.

\para{Domain Similarity embeddings}
We adopt a similar mechanism to extract features of data samples from large pre-trained model as in Domain Similarity \cite{cui_large_2018}.
We aggregate all the representations of the dataset inferred by a probe network as the dataset features. 
The probe network is usually a large network, such as VGG \cite{simonyan_very_2015}, ResNet \cite{he2016identity}.
These networks are pre-trained on \revision{large corpus of data}, e.g., ImageNet \cite{deng_imagenet_2009} \revision{for images or millions of collected texts}, and are considered to be able to capture good generic features from the images and thus serve as reference models to retrieve features. 
The embedding of a dataset $d_k$ is defined as:

\vspace{-3mm}
\begin{equation}
    \tilde{E}_{k} = \sum^{|d_k|}_{j=1} g(x_j), \quad x_j \in d_k,
    \vspace{-0.3cm}
\end{equation}
 $g(\cdot)$ represents the features obtained by extracting the feature layers of the reference model. We adopt \emph{ResNet34} pretrained on ImageNet as the reference model for image datasets \revisiontwo{and \emph{GPT-Neo}~\cite{gpt-neo} for textual datasets}.

\subsubsection{Dataset similarity}
\label{ssec:similarity}
A model with good performance on the source task is likely to have good fine-tuning performance when the target task is similar~\cite{wang_characterizing_2019}.
The similarity between datasets is denoted by $\phi$. 
\revision{This similarity is quantified by calculating the correlation distance between datasets, where a shorter distance signifies greater similarity.}
We expect a higher similarity between semantically similar datasets.
For example, a dataset of flowers shall be more similar to a dataset of plants than airplanes.

\subsection{Other features}



Existing works such as Model2Vec~\cite{achille_task2vec_2019} and attribution map~\cite{song_deep_2019} have investigated to obtain model features for transfer learning. Future work can investigate using model features as an additional type of feature for predicting the model performance.

\section{Graph Construction and Learning}
\label{sec:graph}

The metadata and dataset features mentioned in Section~\ref{sec:features} characterize the datasets and models from a high-level perspective.
When the metadata information and dataset features are similar, distinguishing between them becomes challenging, leading to difficulties in predicting model performance.
In order to obtain more subtle features of models and datasets, we aim to explore the intrinsic relationships between models and datasets. For example, whether a model's proficiency on one dataset implies good performance on a similar dataset, or whether models pre-trained on diverse datasets exhibit distinct performance on a given target dataset. 

\revisiontwo{We introduce a graph-based approach to capture and leverage the relationships between models and datasets. Specifically, we incorporate prior knowledge about dataset-dataset and model-dataset as edges/relationships in the graph. Through graph-based learning, we seek to exploit not only the available node features but also the inherent assumptions or preferences (inductive biases) embedded within the graph's topology.}
The subsequent section will detail how we construct this graph, tailored for model selection with a model zoo.
In addition, we introduce the representative graph learning algorithms that capture the information of the constructed graph.

\subsection{Graph construction}

\begin{figure}[tb]
\centering
	\includegraphics[width=0.38\textwidth]{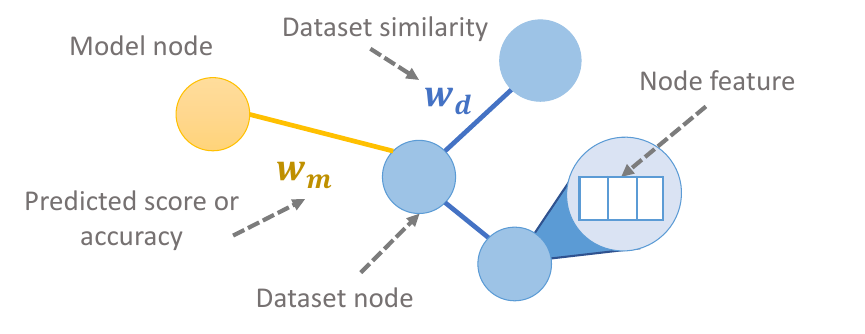}
		\caption{Graph properties}
		\label{fig:graph}
  \vspace{-0.4cm}
\end{figure}

To assign attributes to nodes and edges, it is crucial to identify entities and relationships.  
In Figure~\ref{fig:graph}, we present an overview of the graph structure, where nodes and edges may carry distinct semantic meanings.

\subsubsection{Nodes} A  node in the constructed graph can be either a model or a dataset. The nodes are connected to each other, embedded with model-dataset relationships or dataset-dataset relationships. 
Usually, model zoos contain models trained on overlapping publicly available (benchmark) datasets, making the number of models exceed the number of datasets in a model zoo.

\subsubsection{ Node features}

A  node can be embedded with features.
Some graph learners, e.g., GraphSAGE~\cite{hamilton_inductive_2017}, GAT~\cite{velickovic_graph_2018}, can capture the  node features and use them to initiate the learning process.
We introduce dataset features earlier in Section~\ref{sec:dataset-metadata}. We can embed the dataset features as the features of the dataset nodes.

\subsubsection{Edges and edge attributes}

The edges are constructed in three ways: i) edges between datasets indicating the similarity between datasets, ii) model performance on datasets as edges between models and datasets, iii) predicted scores obtained from other feature-based model selection strategies as another type of edges between models and datasets. 

\para{Dataset-Dataset (D-D) edge attributes} The construction of D-D edges is achieved by evaluating the similarity of dataset representations. The dataset similarity is denoted as $\phi$. The computation encompasses all possible dataset pairs, with the resulting similarity scores employed as edge attributes.

\para{Model-Dataset (M-D) edge attributes} 
    The edges between datasets and models are associated with different meanings. 
    A model can connect to a dataset with training performance or predicted score.
    For example, if we can obtain the pre-trained performance of $m_{Resnet50}$ on the dataset \emph{cifar100} with an accuracy of, e.g., 95\%, the nodes between $m_{Resnet50}$ and \emph{cifar100} has an edge with an attribute of 0.95.
    We can also embed the fine-tuning results if they are available.
    In addition, the predicted scores obtained from other model selection strategies can also embed meaningful information between models and datasets. 
    

\subsection{Graph Learning}

In the context of model selection, we formulate the graph structure to address a link prediction task, evaluating the likelihood of a model exhibiting high performance on the target dataset. The connectivity between the model and dataset nodes is established if the model is anticipated to yield favorable outcomes.

For effective resolution of the link prediction problem, it is imperative to distinguish positive edges from negative ones. In our pursuit of identifying high-performing models, we designate relationships where a model demonstrates good performance on the dataset as positive edges, while those with lower accuracy are categorized as negative edges.

We employ diverse graph learning algorithms for the acquisition of knowledge from the constructed graph. These algorithms consider a variety of information, e.g., link structure and edge attributes. In essence, graph learning algorithms demonstrate the capability to capture intrinsic knowledge within a graph by assimilating neighborhood information.


\subsubsection{Random-walk-based graph learning algorithms}

Graph learning algorithms based on random walks do not incorporate the features of nodes; instead, they focus on learning the graph's link structure. This paper specifically explores Node2Vec~\cite{grover_node2vec_2016} and its variant, Node2Vec+~\cite{liu_accurately_2023}.

\para{Node2Vec}
     Node2Vec~\cite{grover_node2vec_2016} introduces a probability model where the random walk has a certain probability, $1/p$, to revisit nodes being traversed. Additionally, it employs an in-out parameter, $q$, to control the exploration of the global structure. When the return parameter, $p$, is small, the random walk may become trapped in a loop, focusing on the local structure. Conversely, when the in-out parameter, $q$, is small, the random walk resembles a depth-first-sampling strategy more closely, capable of preserving the global structure in the embedding space.
    
\para{Node2Vec+}
    Node2Vec+~\cite{liu_accurately_2023} is a variant of Node2Vec. Different from Node2Vec traversing the graph with parameters, $p$ and $q$, Node2Vec+ takes into account the edge weights. When it constructs walks in the graph, the probability of visiting the next neighbor is associated with the edge weights.

\subsubsection{Neural-network-based learning methods}

Different graph neural networks can learn different kinds of information from the graph. All of them capture the edges in the graph. Some also learn from the edge attributes, or node features. 

\para{GraphSAGE}
GraphSAGE~\cite{hamilton_inductive_2017} employs a sampling and aggregation method to perform inductive node embedding, utilizing node features such as text attributes, node profiles, and more. The model trains a set of aggregation functions that integrate features from the local neighbors and pass them to the target node, denoted as $v_i$. Subsequently, the hidden state of the node $v_i$ is updated by:
    \begin{equation}
        h_i^{(k+1)} = ReLU \left( W^{(k)}h_i^{(k)}, \sum_{n \in N(i)}(ReLU(Q^{(k)}h_n^{(k)})) \right)
    \end{equation} 

\subsubsection{Attention graph embedding}

We also consider another type of graph learning method, using attention mechanisms in the learning process. The attention mechanisms enable graph learning to concentrate on specific parts of a graph that are more relevant to a given task. One advantage of applying attention to graphs is the ability to filter out the noisy components, thereby increasing the signal-to-noise ratio in information processing.
In this line of work, we adopt graph attention networks (GAT) in this paper.

\para{GAT}
GAT~\cite{velickovic_graph_2018} employs masked self-attentional layers to address the limitations of prior graph convolutional-based methods. The layers aim to compute attention coefficients.
\begin{equation}
    \alpha_{ij} = \frac{exp(LeakyReLU(\overrightarrow{a}^{T}[W\overrightarrow{h_i}||W\overrightarrow{h_j}]))}{\sum_{k\in N_i}exp(LeakyReLU(\overrightarrow{a}^{T}[W\overrightarrow{h_i}||W\overrightarrow{h_j}]))},
\end{equation}
$W$ is the weight matrix of the initial linear transformation. The transformed information for each neighbor's feature is then concatenated to derive the new hidden state. This new hidden state undergoes a \emph{LeakyReLU} activation function, a widely utilized rectifier. The attention mechanism described above constitutes a single-layer feed-forward neural network, parameterized by the weight vector mentioned earlier.


\section{The Framework of TransferGraph}
\label{sec:frame}

\begin{figure*}[t]
    \centering
    \includegraphics[width=\linewidth]{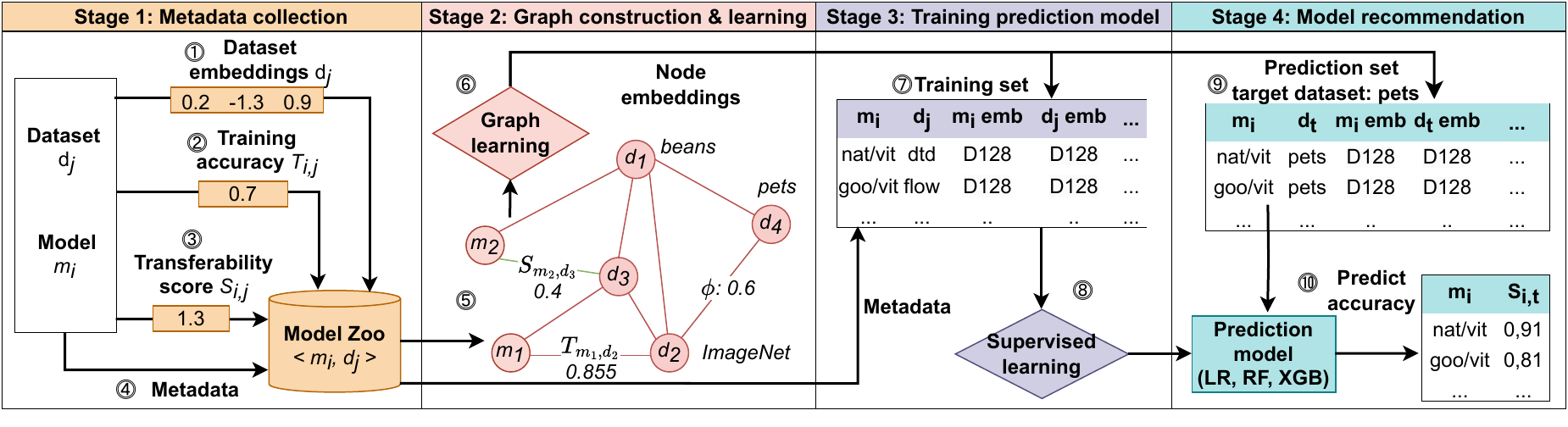}
    \caption{An overview of TransferGraph on model selection for fine-tuning, including model zoo construction (stage 1), training (stage 2-3) and model selection (stage 4). }
    \label{fig:approach}
      \vspace{-0.5cm}
\end{figure*}

We propose \emph{TransferGraph}, a framework that performs model selection via a graph learning process. There are a few steps in the graph-based model selection process, as shown in Figure~\ref{fig:approach}.
The processes are divided into four main steps:

\subsection{Metadata and Feature collection} 

We first collect all the information needed, as described in Section~\ref{sec:features}. Step
\ding{172}-\ding{175} indicate the collection process of different features and metadata used for the subsequent steps. Step \ding{172} obtains the dataset representations, which can be further applied to compute the similarity between datasets. Step \ding{173} encapsulates the training performance of models across different datasets, while step \ding{174} represents the \revision{acquisition of transferability scores of models}, which can be obtained from existing works, e.g., LogME~\cite{you_logme_2021}.
Step \ding{175} collects the metadata of models and datasets.
All the collected information will be returned to the model zoo and stored as preparatory data for further processes.

\subsection{Graph construction and learning}

With the collected information, we continue to construct a graph in step \ding{176}, encapsulating relationships between models and datasets, and other attributes. The graph \revision{component and learning} details are provided in Section~\ref{sec:graph}. 

\revisionthree{We embed different types of relationships in the graph. 
Datasets are connected to each other with edge weights encoding their similarity. Models are connected to datasets with weights of the training performance and/or transferability scores.
To preserve the graph's density and facilitate graph learning, we set specific heuristics during graph construction. These heuristics include setting thresholds to differentiate positive edges from negative ones, based on the edge weight. An positive edge between a model and dataset is established only when the normalized fine-tune accuracy and the transferability score meet or exceed the threshold.
The heuristics and properties of the constructed graph are shown in Table~\ref{tab:graph}.}

\input{sections/graph_property}

    
    

We further use one of the graph learners, e.g., Node2Vec, presented in Section~\ref{sec:graph} to capture the information in the graph, e.g., link structure or  node features, as in step \ding{177}.
The graph learner is trained for a link prediction task. 
With the trained graph learner, we extract the representations for each  node, whose dimension is 128.

\subsection{Training prediction model to predict model performance}

As a learning-based strategy, we learn from the training history to predict the model performance on an unseen dataset as a regression task.
In step \ding{178}, we construct a training set for the supervised learning as a regression task.
The label is the training performance of a model on a dataset.
The training features are constructed by metadata of models and datasets, as well as the  node representations of the models and datasets.
\revisionthree{For example, given the performance of model $m_A$ on dataset $d_B$, we identify the metadata of $m_A$ and $d_B$, as well as the  node representations of them. The information is treated as features and train a prediction model.}


The training set can be represented as tabular data. The prediction models are introduced below:
We then can learn a prediction model, e.g., linear regression, random forest, on the prepared training set, as shown in step \ding{179}.

\para{Linear regression} One of the prediction model we use is linear regression. 
    We use the linear regression model to learn various features, e.g., meta features and graph features.
    Linear regression fits a straight line or surface that minimizes the discrepancies between predicted and actual output values. 

\para{Random forest} Random forest is also a highly adopted model due to its simplicity and explainability. We set the number of trees as 100, max depth as 5.

\para{XGBoost} XGBoost (eXtreme Gradient Boosting) is one of the ensemble learning methods and is particularly effective in structured and tabular data scenarios~\cite{chen2016xgboost}. XGBoost is an ensemble of decision trees and minimizes the objective function with gradient descent. We set the number of trees as 500, and maximum depth as 5.

\subsection{Model recommendation for fine-tuning}

We construct a prediction set \ding{180} similarly to the training set construction.
\revisionthree{Especially, the dataset included in the prediction set is the target dataset we want to predict the model performance on.}
\revisionthree{We adopt a leave-one-out approach for the evaluation of our methodology. When training the prediction model, we utilize all the fine-tuning results from the pairs of models and datasets, excluding the target dataset. In the prediction set, we predict the performance of pairs between all models and the target dataset, i.e., $d_t$.}
The metadata of the dataset also adjust with the target dataset.
We include all the models, since we would like to predict performance of the models in the model zoo on the target dataset.
\revisionthree{More details of the evaluation can be found in Section~\ref{para:evaluation}(Evaluation)}.

Given the trained prediction model, we obtain a score for each model and target dataset pair. We apply these predicted scores as an indicator to rank and select models for fine-tuning.


%% file: sections/graph_property.tex
\begin{table}
    \begin{center}
        \caption{Statistics of the graph properties. (* indicates that the value vary when the dataset and model collection changes)}
    \label{tab:graph}
    \begin{tabular}{ p{48mm}|c|c } 
     \hline
      \textbf{Graph property}\\ 
     \hline\hline
      Modality & image & text \\\hline
      graph type & homogenous & homogenous \\\hline
      Threshold on transferability score for edge pruning  & \multirow{2}{*}{0.5} & \multirow{2}{*}{0.5} \\\hline  
      Threshold on accuracy for edge pruning & 0.5 & 0.5\\\hline
      Threshold of negative edge identification on accuracy & \multirow{2}{*}{0.5} & \multirow{2}{*}{0.5}\\\hline
      Number of nodes & 265 & 188 \\\hline
      Average node degree* & 20.1 & 8.6 \\\hline
      Number of dataset-dataset edge & 5256 & 550 \\\hline
      Number of model-dataset edge with accuracy weight* & \multirow{2}{*}{1753} &  \multirow{2}{*}{918}\\\hline
      Number of model-datset edge with transferability weight* & \multirow{2}{*}{916} & \multirow{2}{*}{419}
     \\\hline
    \end{tabular}
    \end{center}
    \vspace{-4mm}
\end{table}

%% file: sections/4_evaluation.tex
\section{Evaluation}
\label{sec:evaluation}

\subsection{Experiment setups}
\label{ssec:setup}
\begin{table*}[]
\centering
\caption{\revisiontwo{The properties of the target datasets used for evaluation}}
\label{tab:target-tasks}
\resizebox{\textwidth}{!}{%
\begin{tabular}{@{}lccccccccccc@{}}
\toprule
Image dataset & \multicolumn{1}{c}{caltech101~\cite{fei-fei_learning_2004}} & \multicolumn{1}{c}{cifar100~\cite{krizhevsky_learning_2009}} & 
\multicolumn{1}{c}{dtd~\cite{cimpoi_describing_2014}} & \multicolumn{1}{c}{flowers~\cite{nilsback_automated_2008}} &  \multicolumn{1}{c}{pets\cite{ofxford_pets_2022}} & \multicolumn{1}{c}{smallnorb\_elevation\cite{lecun_learning_2004}}  & \multicolumn{1}{c}{stanfordcars~\cite{krause_3d_2013}} & \multicolumn{1}{c}{svhn~\cite{netzer_reading_2011}} \\ \midrule 
Samples & 3060 & 50000  & 1880  & 1020 & 3680 & 24300 & 8144 & 73257 \\ 
Classes & 101 & 100  & 47  & 10  & 37 & 18  & 196 & 10 \\  
\midrule
\midrule
\revisiontwo{Textual dataset} &
  \multicolumn{1}{c}{\revisiontwo{glue/cola\cite{wang_glue_2018}}} &
  \multicolumn{1}{c}{\revisiontwo{glue/sst2\cite{wang_glue_2018}}} &
  \multicolumn{1}{c}{\revisiontwo{rotten\_tomatoes}} &
  \multicolumn{1}{c}{\revisiontwo{tweet\_eval/emotion\cite{barbieri_tweeteval_2020}}} &
  \multicolumn{1}{c}{\revisiontwo{tweet\_eval/hate\cite{barbieri_tweeteval_2020}}} &
  \multicolumn{1}{c}{\revisiontwo{tweet\_eval/irony\cite{barbieri_tweeteval_2020}}} &
  \multicolumn{1}{c}{\revisiontwo{tweet\_eval/offensive\cite{barbieri_tweeteval_2020}}} &
  \multicolumn{1}{c}{\revisiontwo{tweet\_eval/sentiment\cite{barbieri_tweeteval_2020}}} \\ \hline
\revisiontwo{Samples} &
  \multicolumn{1}{c}{\revisiontwo{8550}} &
  \revisiontwo{70000} &
  \revisiontwo{10662} &
  \revisiontwo{5050} &
  \revisiontwo{13000} &
  \revisiontwo{4600} &
  \revisiontwo{24300} &
  \revisiontwo{59900} \\
\revisiontwo{Classes} &
  \revisiontwo{2} &
  \revisiontwo{2} &
  \revisiontwo{2} &
  \revisiontwo{4} &
  \revisiontwo{2} &
  \revisiontwo{2} &
  \revisiontwo{18} &
  \revisiontwo{3} \\

\bottomrule
\end{tabular}%
}

\end{table*}

\para{Datasets}
For evaluation, we collected 12 public image datasets \revisiontwo{and eight textual datasets}, which are often used for classification benchmarks and are listed in Table~\ref{tab:target-tasks}. \revisiontwo{We also included 61 image and 16 textual source datasets, which were used to compute dataset similarity.}



\para{Models}
We include 185 heterogeneous models for image classification tasks \revisiontwo{and 163 models for text classification tasks}, with different architectures, such as ViT~\cite{dosovitskiy_image_2021}, Swin-Transformer~\cite{liu_swin_2021} and ConvNeXT~\cite{liu_convnet_2022} for visual models \revisiontwo{and BERT~\cite{devlin_bert_2019}, FNet~\cite{lee-thorp_fnet_2022} and ELECTRA~\cite{clark_electra_2020} for NLP models}, and pre-trained on diverse datasets.
We use public image \revisiontwo{and text} classification models from HuggingFace\footnote{https://huggingface.co/models}. 
Different from the setup in the previous works~\cite{you_logme_2021,li_guided_2023}, we do not constrain the model to be trained only on a certain dataset, e.g., ImageNet.

 \begin{figure}[t]
    \centering
    \begin{minipage}{0.5\textwidth}
    \centering
        \includegraphics[width=0.88\textwidth]{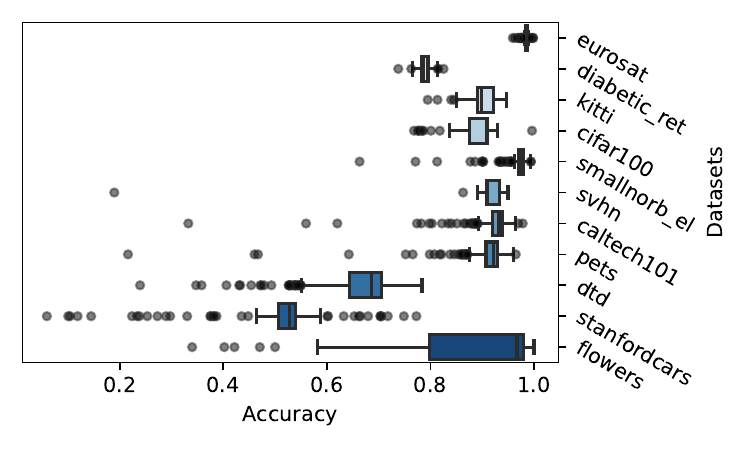}
    \subcaption{Image datasets}
    \end{minipage}
    \centering
    \begin{minipage}{0.5\textwidth}
    \centering
    \includegraphics[width=0.88\textwidth]{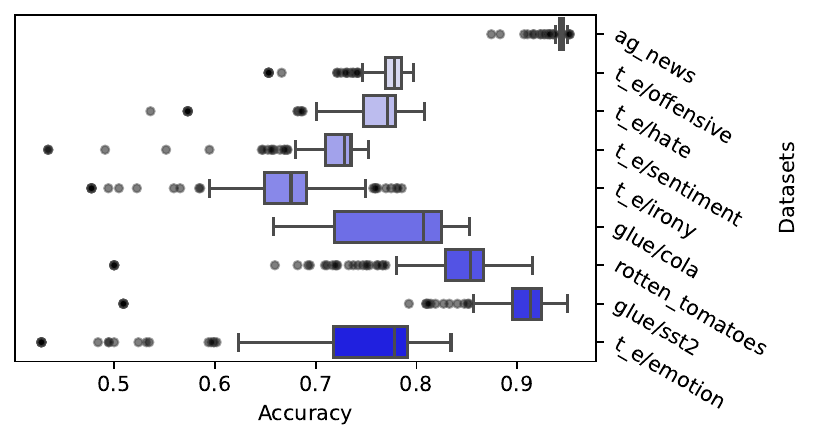}
    \subcaption{\revisiontwo{Textual datasets}}
    \end{minipage}
    \caption{\revisiontwo{Fine-tuning performance of models over different datasets sorted by standard deviation (\textit{t\_e} short for \textit{tweet\_eval})} }
    \label{fig:ft-box}
      \vspace{-0.5cm}
\end{figure}

\para{Ground truth}
A pre-trained deep learning model consists of two components: a \emph{feature extractor} and a \emph{classifier}. 
During fine-tuning, the model is initiated with the pre-trained weights, coupled with a classifier layer that is randomly initialized. 
Subsequently, this new model is retrained on the target dataset.
To determine the actual fine-tuning accuracy, we fine-tune all models on our target datasets, using setups that generalize well over the different target datasets:

\begin{itemize}
    \item For fine-tuning image classification models, we employ \emph{stochastic gradient descent} in combination with a cyclical learning rate scheduler~\cite{smith_cyclical_2017}. We optimize for 30 epochs, using a momentum of 0.9 and max learning rate of 1e-3.

    \item \revisiontwo{For text classification models, we used \emph{AdamW}~\cite{loshchilov_decoupled_2018} in combination with a linear learning rate scheduler and optimized for five epochs, using betas (0.9, 0.999), epsilon 1e-8 and initial learning rate 5e-5.}
\end{itemize}

We present the fine-tuning performance of models across different datasets, as in Figure~\ref{fig:ft-box}. 
Notably, in certain datasets, the performance variance is small. For example, in the case of \emph{eurosat}, where the standard deviation is only 0.005, model selection is not necessary.
In the following experiments, we only report results on datasets where model performance exhibit variation. The datasets are ordered by the standard deviation of the performance.


\para{Baselines}
\label{sssec:baselines}
We compare our work with the two baselines: 

- \texttt{LR} (Amazon LR~\cite{li_guided_2023}) is the state-of-the-art approach for model selection for model zoos.  
    It 
    exploits the meta-features of models and datasets, and uses these features to train a linear regression model to predict the fine-tuned accuracy. The metadata of datasets includes data size, number of classes, etc. The metadata of models consists of the model architecture, input size, pre-trained domain, etc. 
    \revision{\texttt{LR} indicates the strategy including only the metadata as features, \texttt{LR\{all,LogME\}} consisting metadata, dataset similarity and LogME score.}
    
- \texttt{LogME}~\cite{you_logme_2021} is one of the most representative works that measure the transferability of a model to a target dataset. 
    Transferability assesses a model's transfer learning performance to a new task (see Section~\ref{sec:related} for explanations). The mechanism of LogME is to estimate the maximum value of label evidence $p(y|R)$ ($R$ is the representations extracted by a model) given features extracted by pre-trained models.

\para{Evaluation}
\label{para:evaluation}
To validate the effectiveness of our approach, we adopt a ``leave-one-out'' (LOO) mechanism for evaluation. This is a standard setting in related works of model selection, such as \cite{li_guided_2023}. 
At each time, we learn from the training history of models trained on the existing datasets while excluding the target dataset.
When constructing the graph in our proposed method, we remove all the edges of models connected to the target dataset node, i.e., the target dataset, while maintaining the edges between datasets.
Then, with the learned GNN, we identify the node representations of models and the target dataset, and use them as the graph features. 

For baseline comparison, we apply an evaluation metrics: 
\textit{Pearson correlation}. Existing methods for model selection mostly predict a score, i.e., model selection score, for each pair of a model and target dataset. The Pearson correlation measures the correlation between the predicted scores and the ground-truth results, i.e., accuracy. A model selection method is considered better, with a higher correlation between its predicted score and the ground truth.

\para{Summary of our proposed graph-learning-based strategy}
There are a few design choices with our methods. 

\begin{itemize}
    \item \textbf{Prediction model} We include linear regression model (\texttt{LR}), random forest model (\texttt{RF}), XGBoost model (\texttt{XGB}).

    \item \para{Graph learners} The graph learning algorithms include GraphSAGE~\cite{hamilton_inductive_2017}, GAT~\cite{velickovic_graph_2018}, Node2Vec~\cite{grover_node2vec_2016} and Node2Vec+~\cite{liu_accurately_2023}. In particular, \texttt{N2V(+)} is short for  Node2Vec(+) in the plots.

    \item \para{Additional features for supervised learning} Along with the graph features, we also include additional features as inputs for supervised learning when predicting the training results.
    We take into account features including all the metadata of models and datasets, as in Section~\ref{sssec:metadata_model}. In addition, we include the distance between the source dataset and target as another type of features for supervised learning, as in Section~\ref{ssec:similarity}.
\end{itemize}

The variants of our proposed strategies begin with \texttt{TG}. For example, \texttt{TG:LR,N2V,all} indicates that we use a linear regression model \revision{\textit{LR}} to learn from \emph{all} \revision{(including both the basic metadata and dataset similarity)} supervised features along with the graph features obtained by Node2Vec \revision{\textit{N2V}}. 
\revision{\texttt{TG:LR,N2V} includes only the graph features obtained from Node2Vec. }



\subsection{Evaluation on heterogeneous model zoo}

\begin{figure}[tb]
  \vspace{-0.2cm}
    \centering
    \begin{minipage}{0.5\textwidth}
    \centering
    \includegraphics[width=0.75\textwidth]{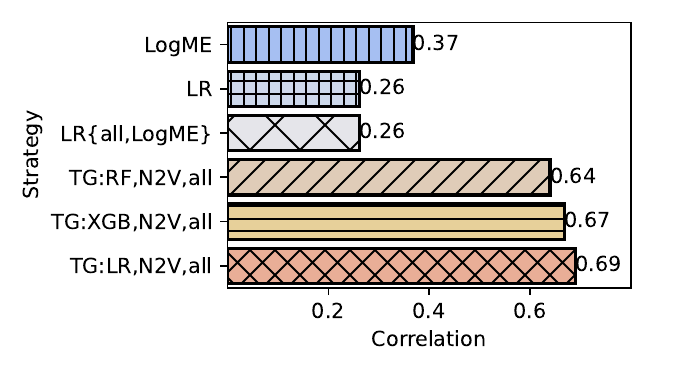}
    \subcaption{Image datasets}
    \label{fig:avg-corr-visual}
    \end{minipage}
    \begin{minipage}{0.5\textwidth}
        \centering\includegraphics[width=0.76\textwidth]{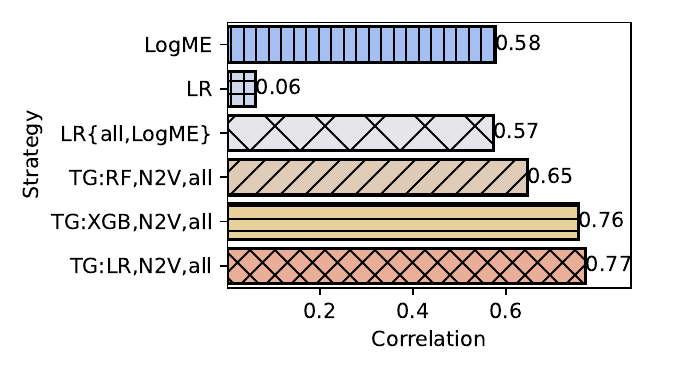}
  \subcaption{\revisiontwo{Textual datasets}}
    \label{fig:avg-corr-text}
    \end{minipage}
    \caption{\revisiontwo{Comparing model selection strategies: feature-based (\texttt{LogME}), learning-based (\texttt{LR}), and our graph-learning-based (\texttt{TG}). Variants \texttt{LR} and \texttt{LR\{all,LogME\}} differ by feature use; \texttt{LR} applies basic metadata, whereas \texttt{LR\{all,LogME\}} includes metadata, dataset distance, and LogME scores. Our \texttt{TG} approaches use different predictive models with metadata, dataset distance, and graph features.}}
    \label{fig:avg_corr}
      \vspace{-0.5cm}
\end{figure}

\revision{We first evaluate the effectiveness of our model selection strategies with the heterogeneous model zoo.}
In Figure~\ref{fig:avg_corr}, we report the average Pearson correlation between the predicted score and the fine-tuning results over \revisiontwo{16} downstream datasets, \revisiontwo{eight for each modality, i.e., text and image}. 
We compare our graph-learning-based strategy with other strategies mentioned in Section~\ref{sssec:baselines}, i.e., \texttt{LogME}, and \texttt{LR}.
\texttt{LogME} is feature-based and does not take into account of meta features nor the source dataset distance.
The rest are all learning-based model selection strategies.
They learn from the training history and predict the model performance on a target dataset.
\revision{\texttt{LR} learns from the basic metadata, while \texttt{LR\{all,LogME\}} also includes the dataset similarity and transferability scores obtained by LogME.}
We present graph-feature-based strategies, beginning with \texttt{TG}. 


Figure~\ref{fig:avg_corr} shows that our proposed graph-feature-based strategies significantly improve the model selection performance compared to baselines \texttt{LogME}, \texttt{LR(\{all,LogME\})}. We use three kinds of prediction models, i.e., linear regression model \texttt{LR}, random forest model \texttt{RF}, and XGBoost model \texttt{XGB}. 
Compared only using (meta) features (\texttt{LR}), the graph features can improve the capability of the prediction model and achieve a higher correlation between the predicted scores and the fine-tuning accuracy. 
\revision{It shows that adding additional information represented in a graph structure can help the prediction of model performance.}
\revisiontwo{We also notice that, on textual datasets, \texttt{LR\{all,LogME\}} significantly outperforms \texttt{LR}, which implies the significance of the selected features.}



\subsection{Ablation study}
\label{ssec:ablation}

\begin{figure}[tb]
    \centering
    \begin{minipage}{0.49\textwidth}
      \includegraphics[width=1\textwidth]{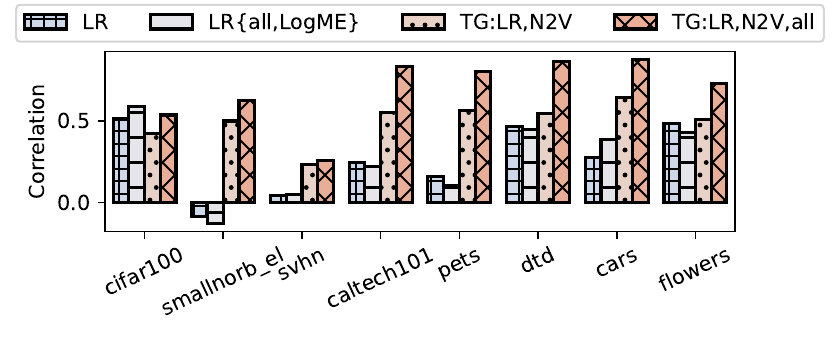}
    \subcaption{Image datasets}
    \label{fig:ablation-visual}  

    \end{minipage}
    \begin{minipage}{0.49\textwidth}
        \includegraphics[width=1\textwidth]{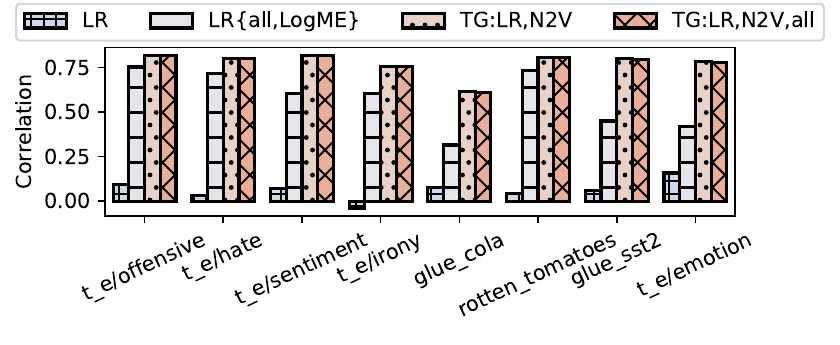}
    \subcaption{\revisiontwo{Textual datasets}}
    \label{fig:ablation-text}
 \end{minipage}
    \caption{\revisiontwo{Ablation study when including various features, i.e., i) with only metadata, ii) with metadata, dataset similarity and LogME scores, iii) with only graph-based features, and iv) with metadata, dataset similarity and graph-features.}}
    \label{fig:ablation}
      \vspace{-0.5cm}
\end{figure}

In this experiment, we conduct an ablation study where we investigate the effect of different features. We use the same prediction model, \texttt{LR} to learn from the features.
As seen from Figure~\ref{fig:ablation}, including more features results in better performance. 
In particular,our approach using graph features significantly outperform the baselines.
We note that when \texttt{LR} fails to learn, e.g., \emph{smallnorb\_evaluation}, the strategies using the graph features can successfully predict the model performance.
Among all, the most effective strategy is to include all the features, i.e., metadata features, dataset similarity, and graph features.
\revisionthree{We also notice that the adding graph features does not yield significant benefits on \textit{cifar100}.
We observe that the performance trends for models on \textit{cifar100} exhibit variability, e.g., models that achieve medium performance on other datasets tend to underperform on \textit{cifar100}. 
Future research could explore strategies to account for this performance variation, using it as prior knowledge to refine the approach.}

\para{\revisionthree{Scenarios without training history}}
\revisionthree{ It is worth noting that we explored scenarios lacking training history on image datasets, which is relevant in initial status. Here, we leverage transferability scores, like LogME, for performance estimation. Despite reduced information, our approach still outperforms baselines, achieving average correlations of 0.47 (with metadata, dataset similarity and graph features) and 0.42 (using only graph features).}

\subsection{Effect of graph learning methods}
In the previous study, we investigate the effect of different features. We move forward to verify the effectiveness of different graph learning methods.
In the following, we compare the average performance using different graph learning algorithms to extract the graph features.
All the strategies included in this experiment learn an \texttt{LR} model to \revision{predict the fine-tuning performance.}
The graph features are extracted by four graph learners: i) \emph{GraphSAGE}~\cite{hamilton_inductive_2017}, ii) \emph{GAT}~\cite{velickovic_graph_2018}, iii) \emph{Node2Vec+}~\cite{liu_accurately_2023}, and iv) \emph{Node2Vec}~\cite{grover_node2vec_2016}.

\begin{figure}[tb]
    \centering
    \includegraphics[width=0.41\textwidth]{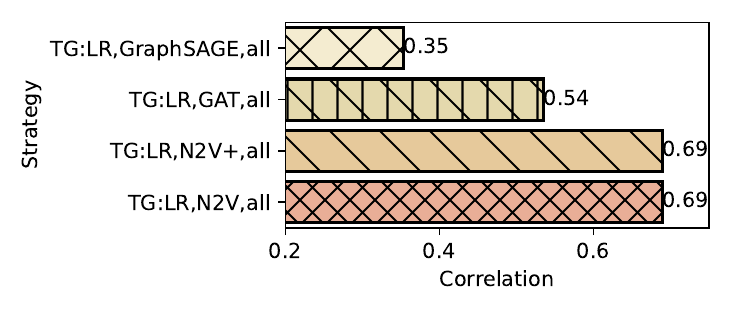}
    \caption{Performance of model selection strategies using different graph learners}
    \label{fig:graph_learner}
      \vspace{-0.5cm}
\end{figure}

Figure~\ref{fig:graph_learner} presents the correlation results when using different graph features obtained by various graph learners. 
The strategies learning features from the Node2Vec series, i.e., \emph{Node2Vec(+)}, outperform the ones using \emph{GraphSAGE} and \emph{GAT}.
Each graph learners consume different graph properties. \emph{Node2Vec} only learns the link structure. Besides the link structure, \emph{Node2Vec+} also takes into account the edge attributes in the graph. While \emph{GraphSAGE} and \emph{GAT} obtain not only the link structure, edge attributes, but also the node features, each updating the node representations in different mechanisms. 

The graph neural networks usually work well on large graphs, e.g., \emph{Citation data} containing 302,424 nodes, and \emph{Reddit} with 232,965 edges~\cite{hamilton_inductive_2017}.
\emph{GraphSAGE} and \emph{GAT} do not perform well in our context because the constructed graph is relatively small compared to those graph datasets. The graph used in this paper has only 265 nodes and thousands of edges. The computation overhead of obtaining such large-scale graph dataset is extremely expensive.
While the \emph{Node2Vec} series of graph learners can perform well on various size of graph dataset.
It is noted that we do not explore the hyperparameter space of these graph learners, e.g., walk length, number of neighbors sampled by each node, window size, etc. 
Complementary work can identify the best hyperparameter candidate for each graph learners, and also investigate which graph learner to apply given different setting scenarios, e.g., graph size, link structure and node/edge features.

\subsection{Effect and capability of prediction model}

\begin{figure}[tb]
    \centering
    \includegraphics[width=0.43\textwidth]{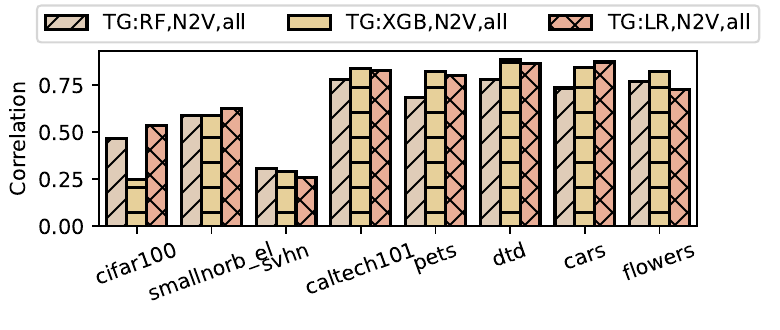}
    \caption{Effect of different prediction models}
    \label{fig:predictor}
       \vspace{-0.5cm}
\end{figure}


The prediction models are used to learn features and predict the fine-tuning scores. In this experiment, we investigate the effect when applying different prediction models. 
In Figure~\ref{fig:predictor}, we present the correlation results when applying different prediction models. We observe that there is no dominant prediction model that can obtain the best results among all the datasets, and the performance on a dataset is similar in general, which indicates that the feature selection is more important than prediction model selection. 
Yet, we do not fully tune the prediction models on each dataset. Further study can be done to identify the most appropriate prediction model based on varying dataset characteristics. 
\vspace{-2mm}
\revisionfour{\subsection{Effect of fine-tuning method}}

\begin{figure}[tb]
    \centering
    \begin{minipage}{0.5\textwidth}
    \centering
    \includegraphics[width=0.75\textwidth]{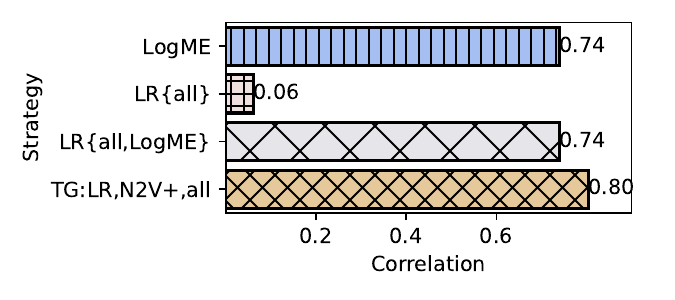}
    \subcaption{\revisionfour{New fine-tuning methods in both training and prediction stage}}
    \label{fig:peft-same}
    \end{minipage}
    \begin{minipage}{0.5\textwidth}
    \centering
    \includegraphics[width=0.75\textwidth]{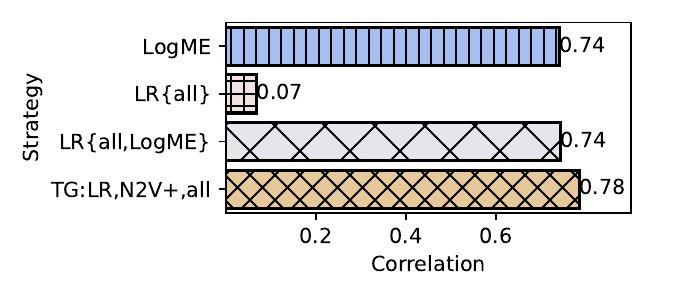}
    \subcaption{\revisionfour{Different fine-tuning methods in training and prediction stage}}
    \label{fig:peft-different}
    \end{minipage}
    \caption{\revisionfour{Comparison of model selection approaches with new fine-tuning method considered}}
    \label{fig:peft}
\end{figure}

\revisionfour{There are multiple current practices to fine-tune models, each yielding different results. In the previous experiments, the fine-tuning method applied retrains all of the layers' pre-trained parameters, which is aiming at effectiveness, but expensive in terms of memory used. 
We adopt another fine-tuning method, \emph{LoRA}~\cite{hu_lora_2021}, which is recently developed aiming at efficiency, both in time and memory. The mechanism is to freeze all model parameters and injects trainable rank decomposition matrices into each layer to reduce the number of trainable parameters. 
This enables the use of larger batch sizes and learning rates, achieves quicker convergence, but may lead to slightly reduced performance. 
We set an initial learning rate of 4e-4 and optimize for four epochs.}

\revisionfour{To investigate the effect of different fine-tuning method, we repeat the experiments for the textual datasets using the new fine-tuning method.}
\revisionfour{Below, in Figure~\ref{fig:peft}, we show the results for both i) repeating the entire experiment with the new fine-tuning results, and ii) keeping the graph constructed with the previous fine-tuning results, but taking the new fine-tuning results as ground truth for the unseen dataset. We show that, for both settings, our graph-based approach consistently outperform the baselines.
Compared to i), using different fine-tuning methods can result in slightly decrease in correlation performance, indicating that different fine-tuning methods do not impact the effectiveness of the approach a lot.}


\subsection{Discussion}
Through comprehensive experiments, we have shown the efficacy of  graph-based features in addressing the model selection problem with a model zoo.
Our most competitive model selection strategy incorporates both graph-based features and additional metadata of models and datasets.
It is noted that, in this paper, we use image classification task and visual models as illustrative scenarios. 
However, our proposed model selection strategy can be applied to diverse cases on various modalities. 
Below, we discuss the limitations and directions that can be investigated in future research.

\para{Graph construction} We incorporate different information in a graph, e.g., dataset distance, model performance, dataset representations, etc. Yet, we do not discuss the contribution and importance of each type of features embedded in a graph. We apply a simple threshold-based edge pruning process to maintain the graph structure. Future work can investigate a more advanced graph construction method and make it adapt to the capability of different graph learning models.

\para{Efficiency} Collecting the relevant features for the prerequisite works is not trivial, though this process can be achieved off-line. Future work can investigate the most impactful features and make the preparation process more efficient.

\para{Graph learning}
We investigate four types of graph learner to obtain graph features. In the graph community, the performance of the graph learner may depend on the graph properties. Further work can pursue to identify good candidates of graph learner (with tuned hyperparameters) for the graph generated from each specific model zoo.
\revisionfour{As future work, dynamic graph learning~\cite{you2022roland} can be investigated for continuous updates. The current approach requires retraining of the graph learner and regression model whenever new nodes or edges are added. 
By dynamically updating the graph learner, we extend TransferGraph to support timely update of the model recommendation.}
\revisionfour{Moreover, future work can adapt methods~\cite{piaggesi2023dine,yuan2022explainability} to interpret and explain the graph learning process to improve the transparency of the model
selection.}



%% file: sections/2_related.tex
\section{Related work}
    
\label{sec:related}

\subsection{Transfer learning}


Traditional machine learning techniques have seen significant progress in various knowledge engineering areas such as classification, regression, clustering and data mining. Despite these advancements, real-world applications frequently encounter limitations.
Unfortunately, in many scenarios, obtaining sufficient and representative training data can be a costly and time-consuming effort. 
Transfer learning has been very successful in combatting these problems, especially in the domain of deep learning, where the \emph{data dependence} is even greater\cite{tan_survey_2018}.

The process of transfer learning typically begins with selecting an \emph{upstream} or \emph{pre-trained} model from a repository containing models trained on different source datasets and architectures. Subsequently, one or multiple selected models are then fine-tuned using the users' target dataset, and the user can select the fine-tuned model with the best characteristics for their downstream task. There are various available fine-tuning strategies identified by\cite{tan_survey_2018}. We adopt the most popular \emph{network-based deep transfer learning} in this work. Network-based deep transfer learning refers to reusing the partial network that pre-trained in the source domain and retraining the deep neural network which used in target domain.  

\subsection{Graph learning}

Graph learning broadly refers to machine learning on data structured as a graph. It is gaining more and more attention, as many complex real-world data can be expressed as graphs. Graph learning can be separated into four different methods: i) \emph{graph signal processing}, ii) \emph{matrix factorization}, iii) \emph{random walk} and iv) \emph{neural network}\cite{xia_graph_2021}.
We focus on the latter two methods, as those are mainly used in graph learning-based recommender systems \cite{wang2021graph}.


\subsubsection{Random-walk-based graph learning algorithms}

These types of algorithms sample random walks by traversing the graph. Given a walk length, i.e., number of steps, a random vertex is selected as the starting point and a neighbor vertex would be selected with probability as the next step in the walk. 
These walks indicate the context of connected vertices. The randomness of walks gives the ability to explore the graph and capture both the global and the local structural information by walking through neighboring vertices.
 After the walks are built, probability models, such as skip-gram~\cite{effic_mikolov_2013}, can be applied to learn the representations. 
The mechanism of the random-walk-based graph learning is aiming to make the representations of connected nodes in the vector space closer to each other while disconnected ones further away. In such a way, the representations capture the graph's intrinsic structure.

\subsubsection{Neural-network-based learning methods}

This line of works were inspired by the success of neural network models, RNNs and CNNs. Graph learning methods using RNNs resemble walks sampled from a graph as words, and use natural language processing models to learn representation of vectors. Another family of neural-network-based methods adopt CNN models. The input can be walks sampled from a graph or the entire graph itself. In this work, we only discuss CNN-based learning methods in this category. Representative works include GraphSAGE~\cite{hamilton_inductive_2017}, GCN~\cite{kipf_semi-supervised_2017}.

%% file: sections/5_conclusion.tex
\section{Conclusion}
\label{sec:conclusion}
We explore the use of a graph-learning-based model selection strategy within the model zoo framework and introduce a comprehensive framework to address the intricate model selection problem. Predicting model performance proves to be challenging, given no dominant model excels across all datasets.
Extensive experiments have shown that effectiveness of leveraging the intrinsic relationships between models and datasets for predicting the model performance. 
The most competitive variant of our model selection strategy gains 32\% increase in measuring the correlation of the predicted model performance and the fine-tuning results.
Furthermore, the graph-learning-based model selection strategy can continuously be improved with more metadata and training history in the model zoo.

\section*{Acknowledgment}
This publication is part of the project Understanding Implicit Dataset Relationships for Machine Learning (with project number VI.Veni.222.439 of the research programme NWO Talent Programme Veni which is (partly) financed by the Dutch Research Council (NWO).  This work was supported by the European Union Horizon Programme call HORIZON-CL4-2022-DATA-01, under Grant 101093164 (ExtremeXP). 

%% file: sections/appendix.tex
\newpage
\appendix
\section{Experiments}
\subsection{Effect of dataset representations}
In this section, we aim to identify the effect of different data representations. Data representations are used to compute the relationship between datasets, i.e., data similarity. 

\para{Task2Vec embeddings}
Task2Vec \cite{achille_task2vec_2019} is another method that we implement to obtain node features. Unlike domain similarity, Task2Vec also takes into account the labels of the dataset and learns embeddings for different tasks with pre-trained networks. The main formula to retrieve the Task2Vec embeddings involves computing the diagonal Fisher Information Matrix of the network filter parameters for a given task: 
\vspace{-2.mm}
\begin{equation}
    \tilde{E}_{k} = F, \quad F = \mathbb{E}_{x,y~\tilde{p}(x)p_w(y|x)[\nabla_w log p_w(y|x) \nabla_w log p_w(y|x)^T]}, 
\end{equation}
where F is the diagonal Fisher Information Matrix (FIM). The Task2Vec embedding can then be obtained by averaging the FIM for all weights in each filter of the probe network. This results in a fixed-size vector representation for each task that captures its complexity and semantic similarity to other tasks. The norm of this embedding correlates with the complexity of the task, while the distance between embeddings captures semantic similarities between tasks.

\para{Experiments exploring the effect of dataset representations}
We extract dataset representations using proposed methods from~\cite{achille_task2vec_2019}. The representations are used in two ways: i) to compute the distance between datasets, and ii) used as initial node features for datasets.
In this experiment, we aim to investigate the effect of different dataset representations. 
The dimension of a Task2vec embedding is 13842, and the one of a domain-similarity embedding is 1024, depending on the extraction layer of the reference model.

\begin{figure}[htb]
    \centering
    \includegraphics[width=0.42\textwidth]{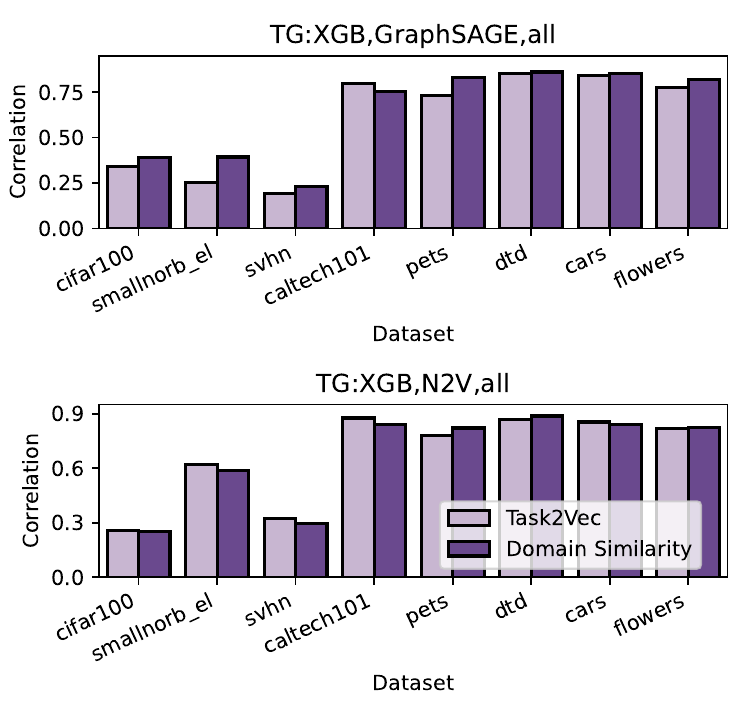}
    \caption{Correlation results affected by different dataset representations.}
    \label{fig:embed}
\end{figure}

In Figure~\ref{fig:embed}, we present the results of two of our proposed strategies, i.e., \texttt{TG:XGB,GraphSAGE,all} and \texttt{TG:XGB,N2V+,all}, using GraphSAGE and Node2Vec+ as graph learner respectively. 
We observe only slight differences in the performance on most of the datasets between using \textit{Task2Vec} representations and the ones of \textit{Domain Similarity}. 
For Nove2Vec+, the embeddings are only used to compute the dataset distance. The small differences in the dataset distance do not affect the final results. 
However, in the case of using GraphSAGE, where the representations are used for both similarity computation and  the vertex features. We observe that in most cases, \textit{Task2Vec} representations do not show advantages when using GraphSAGE. 
One reason is that the \textit{Task2Vec} embeddings have really high dimension, while the graph in general is not big. We suggest that future work can delve into this and identify better representations for a graph learner to learn for the model selection problem.



\subsection{Effect of input ratio}

\begin{figure}[tb]
    \centering
    \includegraphics[width=0.48\textwidth]{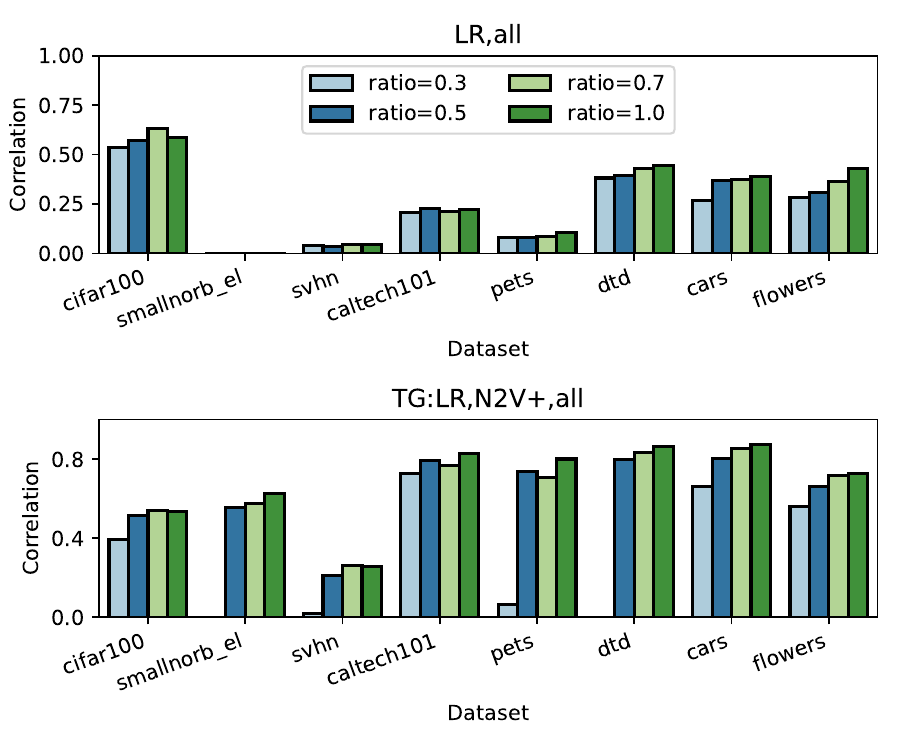}
    \caption{Effect of inputs ratio}
    \label{fig:ratio}
\end{figure}

We investigate the effect of the input size of the training history on the performance. 
The entire training history ($ratio=1.0$) include the training results of all the model and dataset pairs, excluding the target dataset and model pairs. We experiment on training with different ratios of the training history: \{0.3, 0.5, 0.7, 1.0\}. The strategy training would be much more efficient with lower input ratio, because the feature collection can be expensive though it can be performed offline. This experiment aims to answer the question:  whether more training history help the prediction?

We compare two main categories, i.e., a strategy training without graph features (\texttt{LR,all}) and another strategy training with graph features (\texttt{TG:LR,N2V+,all}). 
As in Figure~\ref{fig:ratio}, the performance of both strategy can be affected by the input ratio. 
\texttt{LR,all} is more robust even when limited training history is used to train the strategy. While graph-feature-based strategy is more sensitive to the input ratio, especially with low input ratio.
When we set training history  as $ratio$=0.3, \texttt{TG:LR,N2V+,all} fails to predict the performance. The reason is that with a small input ratio, the constructed graph may have a large number of disconnected components. The graph learner fails to capture the global information by traversing the graph. 

%% file: main.bbl
\begin{thebibliography}{10}
\providecommand{\url}[1]{#1}
\csname url@samestyle\endcsname
\providecommand{\newblock}{\relax}
\providecommand{\bibinfo}[2]{#2}
\providecommand{\BIBentrySTDinterwordspacing}{\spaceskip=0pt\relax}
\providecommand{\BIBentryALTinterwordstretchfactor}{4}
\providecommand{\BIBentryALTinterwordspacing}{\spaceskip=\fontdimen2\font plus
\BIBentryALTinterwordstretchfactor\fontdimen3\font minus \fontdimen4\font\relax}
\providecommand{\BIBforeignlanguage}[2]{{%
\expandafter\ifx\csname l@#1\endcsname\relax
\typeout{** WARNING: IEEEtran.bst: No hyphenation pattern has been}%
\typeout{** loaded for the language `#1'. Using the pattern for}%
\typeout{** the default language instead.}%
\else
\language=\csname l@#1\endcsname
\fi
#2}}
\providecommand{\BIBdecl}{\relax}
\BIBdecl

\bibitem{deng_imagenet_2009}
J.~Deng, W.~Dong, R.~Socher, L.-J. Li, K.~Li, and L.~Fei-Fei, ``{ImageNet}: A large-scale hierarchical image database,'' in \emph{2009 {IEEE} Conference on Computer Vision and Pattern Recognition}, pp. 248--255, {ISSN}: 1063-6919.

\bibitem{deshpande2021linearized}
A.~Deshpande, A.~Achille, A.~Ravichandran, H.~Li, L.~Zancato, C.~Fowlkes, R.~Bhotika, S.~Soatto, and P.~Perona, ``A linearized framework and a new benchmark for model selection for fine-tuning,'' \emph{arXiv preprint arXiv:2102.00084}, 2021.

\bibitem{krause_3d_2013}
J.~Krause, M.~Stark, J.~Deng, and L.~Fei-Fei, ``3d object representations for fine-grained categorization,'' in \emph{2013 {IEEE} International Conference on Computer Vision Workshops}, pp. 554--561.

\bibitem{you_logme_2021}
K.~You, Y.~Liu, J.~Wang, and M.~Long, ``{LogME}: Practical assessment of pre-trained models for transfer learning,'' in \emph{Proceedings of the 38th International Conference on Machine Learning}.\hskip 1em plus 0.5em minus 0.4em\relax {PMLR}, pp. 12\,133--12\,143, {ISSN}: 2640-3498.

\bibitem{tran_transferability_2019}
A.~T. Tran, C.~V. Nguyen, and T.~Hassner, ``Transferability and hardness of supervised classification tasks,'' pp. 1395--1405.

\bibitem{nguyen_leep_2020}
C.~Nguyen, T.~Hassner, M.~Seeger, and C.~Archambeau, ``{LEEP}: A new measure to evaluate transferability of learned representations,'' in \emph{Proceedings of the 37th International Conference on Machine Learning}.\hskip 1em plus 0.5em minus 0.4em\relax {PMLR}, pp. 7294--7305, {ISSN}: 2640-3498.

\bibitem{bolya_scalable_2021}
D.~Bolya, R.~Mittapalli, and J.~Hoffman, ``Scalable diverse model selection for accessible transfer learning,'' in \emph{Advances in Neural Information Processing Systems}, vol.~34.\hskip 1em plus 0.5em minus 0.4em\relax Curran Associates, Inc., pp. 19\,301--19\,312.

\bibitem{huang_frustratingly_2022}
L.-K. Huang, J.~Huang, Y.~Rong, Q.~Yang, and Y.~Wei, ``Frustratingly easy transferability estimation,'' in \emph{Proceedings of the 39th International Conference on Machine Learning}.\hskip 1em plus 0.5em minus 0.4em\relax {PMLR}, pp. 9201--9225, {ISSN}: 2640-3498.

\bibitem{yosinski2014transferable}
J.~Yosinski, J.~Clune, Y.~Bengio, and H.~Lipson, ``How transferable are features in deep neural networks?'' \emph{Advances in neural information processing systems}, vol.~27, 2014.

\bibitem{li_guided_2023}
H.~Li, C.~Fowlkes, H.~Yang, O.~Dabeer, Z.~Tu, and S.~Soatto, ``Guided recommendation for model fine-tuning,'' in \emph{2023 {IEEE}/{CVF} Conference on Computer Vision and Pattern Recognition ({CVPR})}.\hskip 1em plus 0.5em minus 0.4em\relax {IEEE}, pp. 3633--3642.

\bibitem{nargesian2019data}
F.~Nargesian, E.~Zhu, R.~J. Miller, K.~Q. Pu, and P.~C. Arocena, ``Data lake management: challenges and opportunities,'' \emph{Proceedings of the VLDB Endowment}, vol.~12, no.~12, pp. 1986--1989, 2019.

\bibitem{terrizzano2015data}
I.~G. Terrizzano, P.~M. Schwarz, M.~Roth, and J.~E. Colino, ``Data wrangling: The challenging yourney from the wild to the lake.'' in \emph{CIDR}.\hskip 1em plus 0.5em minus 0.4em\relax Asilomar, 2015.

\bibitem{10107808}
R.~Hai, C.~Koutras, C.~Quix, and M.~Jarke, ``Data lakes: A survey of functions and systems,'' \emph{IEEE Transactions on Knowledge and Data Engineering}, vol.~35, no.~12, pp. 12\,571--12\,590, 2023.

\bibitem{DBLP:conf/icde/FernandezAKYMS18}
R.~C. Fernandez, Z.~Abedjan, F.~Koko, G.~Yuan, S.~Madden, and M.~Stonebraker, ``{Aurum: {A} Data Discovery System},'' in \emph{{ICDE}}, 2018, pp. 1001--1012.

\bibitem{zhang2020finding}
Y.~Zhang and Z.~G. Ives, ``{Finding Related Tables in Data Lakes for Interactive Data Science},'' in \emph{{SIGMOD}}, 2020, pp. 1951--1966.

\bibitem{nargesian2020organizing}
F.~Nargesian, K.~Q. Pu, E.~Zhu, B.~{Ghadiri Bashardoost}, and R.~J. Miller, ``{Organizing Data Lakes for Navigation},'' in \emph{{SIGMOD}}, 2020, pp. 1939--1950.

\bibitem{renggli_shift_2022}
C.~Renggli, X.~Yao, L.~Kolar, L.~Rimanic, A.~Klimovic, and C.~Zhang, ``{SHiFT}: an efficient, flexible search engine for transfer learning,'' vol.~16, no.~2, pp. 304--316.

\bibitem{wang_characterizing_2019}
Z.~Wang, Z.~Dai, B.~Poczos, and J.~Carbonell, ``Characterizing and avoiding negative transfer,'' in \emph{2019 {IEEE}/{CVF} Conference on Computer Vision and Pattern Recognition ({CVPR})}.\hskip 1em plus 0.5em minus 0.4em\relax {IEEE}, pp. 11\,285--11\,294.

\bibitem{cui_large_2018}
Y.~Cui, Y.~Song, C.~Sun, A.~Howard, and S.~Belongie, ``Large scale fine-grained categorization and domain-specific transfer learning,'' pp. 4109--4118.

\bibitem{achille_task2vec_2019}
A.~Achille, M.~Lam, R.~Tewari, A.~Ravichandran, S.~Maji, C.~C. Fowlkes, S.~Soatto, and P.~Perona, ``Task2vec: Task embedding for meta-learning,'' pp. 6430--6439.

\bibitem{deshpande_linearized_2021}
A.~Deshpande, A.~Achille, A.~Ravichandran, H.~Li, L.~Zancato, C.~C. Fowlkes, R.~Bhotika, S.~Soatto, and P.~Perona, ``A linearized framework and a new benchmark for model selection for fine-tuning.''

\bibitem{he2016identity}
K.~He, X.~Zhang, S.~Ren, and J.~Sun, ``Identity mappings in deep residual networks,'' in \emph{Computer Vision--ECCV 2016: 14th European Conference, Amsterdam, The Netherlands, October 11--14, 2016, Proceedings, Part IV 14}.\hskip 1em plus 0.5em minus 0.4em\relax Springer, 2016, pp. 630--645.

\bibitem{dosovitskiy_image_2021}
A.~Dosovitskiy, L.~Beyer, A.~Kolesnikov, D.~Weissenborn, X.~Zhai, T.~Unterthiner, M.~Dehghani, M.~Minderer, G.~Heigold, S.~Gelly, J.~Uszkoreit, and N.~Houlsby, ``An image is worth 16x16 words: Transformers for image recognition at scale,'' in \emph{9th International Conference on Learning Representations, {ICLR} 2021, Virtual Event, Austria, May 3-7, 2021}.\hskip 1em plus 0.5em minus 0.4em\relax {OpenReview}.net.

\bibitem{lecun_gradient-based_1998}
Y.~Lecun, L.~Bottou, Y.~Bengio, and P.~Haffner, ``Gradient-based learning applied to document recognition,'' vol.~86, no.~11, pp. 2278--2324, conference Name: Proceedings of the {IEEE}.

\bibitem{zamir_taskonomy_2018}
A.~R. Zamir, A.~Sax, W.~Shen, L.~Guibas, J.~Malik, and S.~Savarese, ``Taskonomy: Disentangling task transfer learning,'' in \emph{2018 {IEEE}/{CVF} Conference on Computer Vision and Pattern Recognition}.\hskip 1em plus 0.5em minus 0.4em\relax {IEEE}, pp. 3712--3722.

\bibitem{simonyan_very_2015}
K.~Simonyan and A.~Zisserman, ``Very deep convolutional networks for large-scale image recognition,'' in \emph{International Conference on Learning Representations}, 2015.

\bibitem{gpt-neo}
S.~Black, G.~Leo, P.~Wang, C.~Leahy, and S.~Biderman, ``{GPT-Neo: Large Scale Autoregressive Language Modeling with Mesh-Tensorflow},'' Mar. 2021, {If you use this software, please cite it using these metadata.}

\bibitem{song_deep_2019}
J.~Song, Y.~Chen, X.~Wang, C.~Shen, and M.~Song, ``Deep model transferability from attribution maps,'' in \emph{Advances in Neural Information Processing Systems}, vol.~32.\hskip 1em plus 0.5em minus 0.4em\relax Curran Associates, Inc.

\bibitem{hamilton_inductive_2017}
W.~L. Hamilton, R.~Ying, and J.~Leskovec, ``Inductive representation learning on large graphs,'' in \emph{Proceedings of the 31st International Conference on Neural Information Processing Systems}, ser. {NIPS}'17.\hskip 1em plus 0.5em minus 0.4em\relax Curran Associates Inc., pp. 1025--1035.

\bibitem{velickovic_graph_2018}
P.~Veličković, G.~Cucurull, A.~Casanova, A.~Romero, P.~Liò, and Y.~Bengio, ``Graph attention networks.''

\bibitem{grover_node2vec_2016}
A.~Grover and J.~Leskovec, ``node2vec: Scalable feature learning for networks,'' in \emph{Proceedings of the 22nd {ACM} {SIGKDD} International Conference on Knowledge Discovery and Data Mining}, ser. {KDD} '16.\hskip 1em plus 0.5em minus 0.4em\relax Association for Computing Machinery, pp. 855--864.

\bibitem{liu_accurately_2023}
R.~Liu, M.~Hirn, and A.~Krishnan, ``Accurately modeling biased random walks on weighted networks using node2vec+,'' vol.~39, no.~1, p. btad047, publisher: Oxford University Press.

\bibitem{chen2016xgboost}
T.~Chen and C.~Guestrin, ``Xgboost: A scalable tree boosting system,'' in \emph{Proceedings of the 22nd acm sigkdd international conference on knowledge discovery and data mining}, 2016, pp. 785--794.

\bibitem{fei-fei_learning_2004}
L.~Fei-Fei, R.~Fergus, and P.~Perona, ``Learning generative visual models from few training examples: An incremental bayesian approach tested on 101 object categories,'' in \emph{2004 Conference on Computer Vision and Pattern Recognition Workshop}, pp. 178--178.

\bibitem{krizhevsky_learning_2009}
A.~Krizhevsky, ``Learning multiple layers of features from tiny images.''

\bibitem{cimpoi_describing_2014}
M.~Cimpoi, S.~Maji, I.~Kokkinos, S.~Mohamed, and A.~Vedaldi, ``Describing textures in the wild,'' in \emph{2014 {IEEE} Conference on Computer Vision and Pattern Recognition}.\hskip 1em plus 0.5em minus 0.4em\relax {IEEE}, pp. 3606--3613.

\bibitem{nilsback_automated_2008}
M.-E. Nilsback and A.~Zisserman, ``Automated flower classification over a large number of classes,'' in \emph{2008 Sixth Indian Conference on Computer Vision, Graphics \& Image Processing}.\hskip 1em plus 0.5em minus 0.4em\relax {IEEE}, pp. 722--729.

\bibitem{ofxford_pets_2022}
H.~Zhang, S.~Zhou, G.~Y. Li, and N.~Xiu, ``0/1 deep neural networks via block coordinate descent,'' \emph{CoRR}, vol. abs/2206.09379, 2022.

\bibitem{lecun_learning_2004}
Y.~LeCun, {Fu Jie Huang}, and L.~Bottou, ``Learning methods for generic object recognition with invariance to pose and lighting,'' in \emph{Proceedings of the 2004 {IEEE} Computer Society Conference on Computer Vision and Pattern Recognition, 2004. {CVPR} 2004.}, vol.~2.\hskip 1em plus 0.5em minus 0.4em\relax {IEEE}, pp. 97--104.

\bibitem{netzer_reading_2011}
Y.~Netzer, T.~Wang, A.~Coates, A.~Bissacco, B.~Wu, and A.~Ng, ``Reading digits in natural images with unsupervised feature learning.''

\bibitem{wang_glue_2018}
A.~Wang, A.~Singh, J.~Michael, F.~Hill, O.~Levy, and S.~Bowman, ``{GLUE}: A multi-task benchmark and analysis platform for natural language understanding,'' in \emph{Proceedings of the 2018 {EMNLP} Workshop {BlackboxNLP}: Analyzing and Interpreting Neural Networks for {NLP}}, T.~Linzen, G.~Chrupa\{{\textbackslash}textbackslash\}la, and A.~Alishahi, Eds.\hskip 1em plus 0.5em minus 0.4em\relax Association for Computational Linguistics, pp. 353--355.

\bibitem{barbieri_tweeteval_2020}
F.~Barbieri, J.~Camacho-Collados, L.~Espinosa~Anke, and L.~Neves, ``{TweetEval}: Unified benchmark and comparative evaluation for tweet classification,'' in \emph{Findings of the Association for Computational Linguistics: {EMNLP} 2020}, T.~Cohn, Y.~He, and Y.~Liu, Eds.\hskip 1em plus 0.5em minus 0.4em\relax Association for Computational Linguistics, pp. 1644--1650.

\bibitem{liu_swin_2021}
Z.~Liu, Y.~Lin, Y.~Cao, H.~Hu, Y.~Wei, Z.~Zhang, S.~Lin, and B.~Guo, ``Swin transformer: Hierarchical vision transformer using shifted windows,'' in \emph{2021 {IEEE}/{CVF} International Conference on Computer Vision ({ICCV})}.\hskip 1em plus 0.5em minus 0.4em\relax {IEEE}, pp. 9992--10\,002.

\bibitem{liu_convnet_2022}
Z.~Liu, H.~Mao, C.-Y. Wu, C.~Feichtenhofer, T.~Darrell, and S.~Xie, ``A {ConvNet} for the 2020s,'' in \emph{2022 {IEEE}/{CVF} Conference on Computer Vision and Pattern Recognition ({CVPR})}.\hskip 1em plus 0.5em minus 0.4em\relax {IEEE}, pp. 11\,966--11\,976.

\bibitem{devlin_bert_2019}
J.~Devlin, M.-W. Chang, K.~Lee, and K.~Toutanova, ``{BERT}: Pre-training of deep bidirectional transformers for language understanding,'' in \emph{Proceedings of the 2019 Conference of the North American Chapter of the Association for Computational Linguistics: Human Language Technologies, Volume 1 (Long and Short Papers)}, J.~Burstein, C.~Doran, and T.~Solorio, Eds.\hskip 1em plus 0.5em minus 0.4em\relax Association for Computational Linguistics, pp. 4171--4186.

\bibitem{lee-thorp_fnet_2022}
J.~Lee-Thorp, J.~Ainslie, I.~Eckstein, and S.~Ontanon, ``{FNet}: Mixing tokens with fourier transforms,'' in \emph{Proceedings of the 2022 Conference of the North American Chapter of the Association for Computational Linguistics: Human Language Technologies}, M.~Carpuat, M.-C. de~Marneffe, and I.~V. Meza~Ruiz, Eds.\hskip 1em plus 0.5em minus 0.4em\relax Association for Computational Linguistics, pp. 4296--4313.

\bibitem{clark_electra_2020}
K.~Clark, M.-T. Luong, and Q.~V. Le, ``{ELECTRA}: {PRE}-{TRAINING} {TEXT} {ENCODERS} {AS} {DISCRIMINATORS} {RATHER} {THAN} {GENERATORS}.''

\bibitem{smith_cyclical_2017}
L.~N. Smith, ``Cyclical learning rates for training neural networks,'' in \emph{2017 {IEEE} Winter Conference on Applications of Computer Vision ({WACV})}.\hskip 1em plus 0.5em minus 0.4em\relax {IEEE}, pp. 464--472.

\bibitem{loshchilov_decoupled_2018}
I.~Loshchilov and F.~Hutter, ``Decoupled weight decay regularization.''

\bibitem{hu_lora_2021}
E.~J. Hu, Y.~Shen, P.~Wallis, Z.~Allen-Zhu, Y.~Li, S.~Wang, L.~Wang, and W.~Chen, ``{LoRA}: Low-rank adaptation of large language models.''

\bibitem{you2022roland}
J.~You, T.~Du, and J.~Leskovec, ``Roland: graph learning framework for dynamic graphs,'' in \emph{Proceedings of the 28th ACM SIGKDD Conference on Knowledge Discovery and Data Mining}, 2022, pp. 2358--2366.

\bibitem{piaggesi2023dine}
S.~Piaggesi, M.~Khosla, A.~Panisson, and A.~Anand, ``Dine: Dimensional interpretability of node embeddings,'' \emph{arXiv preprint arXiv:2310.01162}, 2023.

\bibitem{yuan2022explainability}
H.~Yuan, H.~Yu, S.~Gui, and S.~Ji, ``Explainability in graph neural networks: A taxonomic survey,'' \emph{IEEE transactions on pattern analysis and machine intelligence}, vol.~45, no.~5, pp. 5782--5799, 2022.

\bibitem{tan_survey_2018}
C.~Tan, F.~Sun, T.~Kong, W.~Zhang, C.~Yang, and C.~Liu, ``A survey on deep transfer learning,'' in \emph{Artificial Neural Networks and Machine Learning – {ICANN} 2018}, ser. Lecture Notes in Computer Science, V.~Kůrková, Y.~Manolopoulos, B.~Hammer, L.~Iliadis, and I.~Maglogiannis, Eds.\hskip 1em plus 0.5em minus 0.4em\relax Springer International Publishing, pp. 270--279.

\bibitem{xia_graph_2021}
F.~Xia, K.~Sun, S.~Yu, A.~Aziz, L.~Wan, S.~Pan, and H.~Liu, ``Graph learning: A survey,'' vol.~2, no.~2, pp. 109--127.

\bibitem{wang2021graph}
S.~Wang, L.~Hu, Y.~Wang, X.~He, Q.~Z. Sheng, M.~A. Orgun, L.~Cao, F.~Ricci, and S.~Y. Philip, ``Graph learning based recommender systems: A review,'' in \emph{IJCAI International Joint Conference on Artificial Intelligence}.\hskip 1em plus 0.5em minus 0.4em\relax International Joint Conferences on Artificial Intelligence, 2021, pp. 4644--4652.

\bibitem{effic_mikolov_2013}
T.~Mikolov, K.~Chen, G.~Corrado, and J.~Dean, ``Efficient estimation of word representations in vector space,'' in \emph{1st International Conference on Learning Representations, {ICLR} 2013, Scottsdale, Arizona, USA, May 2-4, 2013, Workshop Track Proceedings}, Y.~Bengio and Y.~LeCun, Eds., 2013.

\bibitem{kipf_semi-supervised_2017}
T.~N. Kipf and M.~Welling, ``Semi-supervised classification with graph convolutional networks.''

\end{thebibliography}
